\documentclass[english]{article}

\pdfoutput=1
\usepackage[T1]{fontenc}
\usepackage[latin9]{inputenc}
\usepackage{geometry}
\usepackage{verbatim}
\usepackage{listings}
\usepackage{float}
\usepackage{bm}
\usepackage{bbm}
\usepackage{amsmath}
\usepackage{subcaption}
\usepackage{amssymb}
\usepackage{graphicx}
\usepackage{hyperref}
\usepackage{hhline}
\hypersetup{
 colorlinks,linkcolor=red,anchorcolor=blue,citecolor=blue}
\usepackage{breakurl}
\usepackage{amsthm}
\usepackage{dsfont}
\usepackage{array}
\usepackage{color}

\definecolor{codegreen}{rgb}{0,0.6,0}
\definecolor{codegray}{rgb}{0.5,0.5,0.5}
\definecolor{codepurple}{rgb}{0.58,0,0.82}
\definecolor{backcolour}{rgb}{0.95,0.95,0.92}

\makeatletter

\providecommand{\tabularnewline}{\\}
\floatstyle{ruled}
\newfloat{algorithm}{tbp}{loa}
\providecommand{\algorithmname}{Algorithm}
\floatname{algorithm}{\protect\algorithmname}

\definecolor{yxc}{RGB}{255,0,0}
\definecolor{yjc}{RGB}{125,0,0}
\definecolor{cm}{RGB}{0,0,200}
\definecolor{kzw}{RGB}{0,150,0}

\allowdisplaybreaks

\usepackage{enumitem}
\setlist[itemize]{leftmargin=1em}
\setlist[enumerate]{leftmargin=1em}

\usepackage{algorithm}
\usepackage{algorithmic}
\usepackage{arydshln}

\newcommand{\citep}{\cite}
\newcommand{\citealp}{\cite}
\makeatother

\makeatletter
\newcommand*{\rom}[1]{\expandafter\@slowromancap\romannumeral #1@}
\makeatother

\newcommand{\rank}{\mathrm{rank}}

\newcommand{\var}{\mathrm{Var}}

\newcommand{\vect}{\mathrm{Vec}}

\newcommand{\argmin}{\mbox{argmin}}

\newcommand{\where}{\text{where}}

\newcommand{\btanh}{\mathrm{\textbf{tanh}}}

\newcommand{\R}{\mathbb{R}}

\newcommand{\E}{\mathbb{E}}

\renewcommand{\P}{\mathbb{P}}

\newcommand{\bbone}{\mathbbm{1}}

\newcommand{\bx}{\mathrm{\bf x}}

\newcommand{\bF}{\mathrm{\bf F}}

\newcommand{\bW}{\mathrm{\bf W}}

\newcommand{\bI}{\mathrm{\bf I}}
\newcommand{\bg}{\mathrm{\bf g}}

\newcommand{\bX}{\mathrm{\bf X}}

\newcommand{\xx}{\text{\boldmath $x$}}

\newcommand{\zz}{\text{\boldmath $z$}}

\newcommand{\ww}{\text{\boldmath $w$}}
\newcommand{\kk}{\text{\boldmath $k$}}

\newcommand{\uu}{\text{\boldmath $u$}}

\newcommand{\hh}{\text{\boldmath $h$}}

\newcommand{\bb}{\text{\boldmath $b$}}
\newcommand{\cc}{\text{\boldmath $c$}}
\newcommand{\pp}{\text{\boldmath $p$}}
\newcommand{\bgg}{\text{\boldmath $g$}}
\newcommand{\oo}{\text{\boldmath $o$}}
\newcommand{\ii}{\text{\boldmath $i$}}
\newcommand{\ff}{\text{\boldmath $f$}}

\newcommand{\bbeta}{\text{\boldmath $\beta$}}

\newcommand{\btheta}{\text{\boldmath $\theta$}}
\newcommand{\bomega}{\text{\boldmath $\omega$}}
\newcommand{\bsigma}{\bm{\sigma}}

\newcommand{\bzero}{\mathrm{\bf 0}}

\newcommand{\bxi}{\text{\boldmath $\xi$}}

\newcommand{\cB}{\mathcal{B}}

\newcommand{\cL}{\mathcal{L}}

\newcommand{\cE}{\mathcal{E}}
\newcommand{\cT}{\mathcal{T}}
\newcommand{\cF}{\mathcal{F}}

\newcommand{\cH}{\mathcal{H}}

\DeclareFixedFont{\ttb}{T1}{txtt}{bx}{n}{10} 
\DeclareFixedFont{\ttm}{T1}{txtt}{m}{n}{10}  

\usepackage{color}
\definecolor{deepblue}{rgb}{0,0,0.5}
\definecolor{deepred}{rgb}{0.6,0,0}
\definecolor{deepgreen}{rgb}{0,0.5,0}

\begin{document}
\theoremstyle{plain} 
\newtheorem{lem}{\textbf{Lemma}} 
\newtheorem{prop}{\textbf{Proposition}}
\newtheorem{thm}{\textbf{Theorem}}\setcounter{thm}{0}
\newtheorem{corollary}{\textbf{Corollary}} 
\newtheorem{example}{\textbf{Example}}
\newtheorem{definition}{\textbf{Definition}} 
\newtheorem{fact}{\textbf{Fact}}
\newtheorem{claim}{\textbf{Claim}}

\theoremstyle{definition}

\theoremstyle{remark}\newtheorem{remark}{\textbf{Remark}}\newtheorem{conjecture}{Conjecture}\newtheorem{condition}{\textbf{Condition}}\newtheorem{assumption}{\textbf{Assumption}}

\title{A Selective Overview of Deep Learning\footnotetext{Author
names are sorted alphabetically.}}

\author{Jianqing Fan\thanks{Department of Operations Research and Financial Engineering, Princeton
University, Princeton, NJ 08544, USA; Email: \texttt{\{jqfan, congm,
yiqiaoz\}@princeton.edu}.} \and Cong Ma\footnotemark[3] \and Yiqiao Zhong\footnotemark[1] }

\maketitle
\begin{abstract}
Deep learning has arguably achieved tremendous success in recent years. In simple words, deep learning uses the composition of many nonlinear functions to model the complex dependency between input features and labels. While neural networks have a long history, recent advances have greatly improved their performance in computer vision, natural language processing, etc. From the statistical and scientific perspective, it is natural to ask: What is deep learning? What are the new characteristics of deep learning, compared with classical methods? What are the theoretical foundations of deep learning?

To answer these questions, we introduce common neural network models (e.g., convolutional neural nets, recurrent neural nets, generative adversarial nets) and training techniques (e.g., stochastic gradient descent, dropout, batch normalization) from a statistical point of view. Along the way, we highlight new characteristics of deep learning (including depth and over-parametrization) and explain their practical and theoretical benefits. We also sample recent results on theories of deep learning, many of which are only suggestive. While a complete understanding of deep learning remains elusive, we hope that our perspectives and discussions serve as a stimulus for new statistical research.
\end{abstract}



\setcounter{tocdepth}{2}
\tableofcontents{}


\section{Introduction}\label{sec:intro}
Modern machine learning and statistics deal with the problem of \emph{learning from data}: given a training dataset $\{(y_i,\xx_i)\}_{1\leq i \leq n}$ where $\xx_i \in \mathbb{R}^{d}$ is the input and $y_i \in \mathbb{R}$ is the output\footnote{When the label $y$ is given, this problem is often known as \emph{supervised learning}. We mainly focus on this paradigm throughout this paper and remark sparingly on its counterpart, \emph{unsupervised learning}, where $y$ is not given.}, one seeks a function $f: \mathbb{R}^{d} \mapsto \mathbb{R}$ from a certain function class $\mathcal{F}$ that has good prediction performance on test data. This problem is of fundamental significance and finds applications in numerous scenarios. For instance, in image recognition, the input $\xx$ (reps.~the output $y$) corresponds to the raw image (reps.~its category) and the goal is to find a mapping $f(\cdot)$ that can classify future images accurately. Decades of research efforts in statistical machine learning have been devoted to developing methods to find $f(\cdot)$ efficiently with provable guarantees. Prominent examples include linear classifiers (e.g., linear$\,$/$\,$logistic regression, linear discriminant analysis), kernel methods (e.g., support vector machines), tree-based methods (e.g., decision trees, random forests), nonparametric regression (e.g., nearest neighbors, local kernel smoothing), etc. Roughly speaking, each aforementioned method corresponds to a different function class $\mathcal{F}$ from which the final classifier $f(\cdot)$ is chosen.

Deep learning~\citep{lecun2015deep}, in its simplest form, proposes the following \emph{compositional} function class:
\begin{equation}\label{model:1}
\left\{f(\xx; \btheta) = \bW_{L} \bsigma_L(\bW_{L-1}\cdots \bsigma_{2}(\bW_2 \bsigma_1(\bW_1 \xx))) \;\big| \;\btheta = \{\bW_1,\ldots, \bW_{L}\}\right\}.
\end{equation}
Here, for each $1\leq l \leq L$, $\bsigma_\ell(\cdot)$ is some nonlinear function, and $\btheta = \{\bW_1,\ldots, \bW_{L}\}$ consists of matrices with appropriate sizes. Though simple, deep learning has made significant progress towards addressing the problem of learning from data over the past decade. Specifically, it has performed close to or better than humans in various important tasks in artificial intelligence, including image recognition~\citep{he2016deep}, game playing~\citep{silver2017mastering}, and machine translation~\citep{wu2016google}. Owing to its great promise, the impact of deep learning is also growing rapidly in areas beyond artificial intelligence; examples include statistics~\citep{bauer2017deep, schmidt2017nonparametric, liang2017well, romano2018deep,gao2018robust}, applied mathematics~\citep{weinan2017deep, chen2018neural}, clinical research~\citep{de2018clinically}, etc.

\begin{table}[htb]
\caption{Winning models for ILSVRC image classification challenge.}
\label{tab:intro}
\begin{center}
\begin{tabular}{|c|c|c|c|c|}
\hline
Model & Year & \# Layers & \# Params & Top-5 error \\
\hline
Shallow & $<2012$ & --- & --- & $>25\%$ \\
\hline
AlexNet & $2012$ & $8$ & $61$M & $16.4\%$ \\
\hline
VGG19 & $2014$ & $19$ & $144$M & $7.3\%$ \\
\hline
GoogleNet & $2014$ & $22$ & $7$M & $6.7\%$ \\
\hline
ResNet-$152$ & $2015$ & $152$ & $60$M & $3.6\%$ \\
\hline
\end{tabular}
\end{center}
\end{table}
To get a better idea of the success of deep learning, let us take the ImageNet Challenge~\citep{ILSVRC15} (also known as ILSVRC) as an example. In the classification task, one is given a training dataset consisting of 1.2 million color images with $1000$ categories, and the goal is to classify images based on the input pixels. The performance of a classifier is then evaluated on a test dataset of 100 thousand images, and in the end the top-5 error\footnote{The algorithm makes an error if the true label is not contained in the $5$ predictions made by the algorithm.} is reported. Table~\ref{tab:intro} highlights a few popular models and their corresponding performance. As can be seen, deep learning models (the second to the last rows) have a clear edge over shallow models (the first row) that fit linear models$\,$/$\,$tree-based models on handcrafted features. This significant improvement raises a foundational question:

\begin{itemize}
\centering
\item[] \mbox{\emph{Why is deep learning better than classical methods on tasks like image recognition?}}
\end{itemize}

\begin{figure}
\centering
\includegraphics[width = 0.75\textwidth]{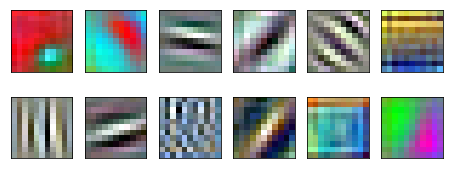}
\caption{Visualization of trained filters in the first layer of AlexNet. The model is pre-trained on ImageNet and is downloadable via PyTorch package \texttt{torchvision.models}. Each filter contains $11 \times 11 \times 3$ parameters and is shown as an RGB color map of size $11 \times 11$.}\label{fig:vis}
\end{figure}
\subsection{Intriguing new characteristics of deep learning}

It is widely acknowledged that two indispensable factors contribute to the success of deep learning, namely (1) huge datasets that often contain millions of samples and (2) immense computing power resulting from clusters of graphics processing units (GPUs). Admittedly, these resources are only recently available: the latter allows to train larger neural networks which reduces biases and the former enables variance reduction. However, these two alone are not sufficient to explain the mystery of deep learning due to some of its ``dreadful'' characteristics: (1) \emph{over-parametrization}: the number of parameters in state-of-the-art deep learning models is often much larger than the sample size (see Table~\ref{tab:intro}), which gives them the potential to overfit the training data, and (2) \emph{nonconvexity}: even with the help of GPUs, training deep learning models is still NP-hard~\citep{arora2009computational} in the worst case due to the highly {nonconvex} loss function to minimize. In reality, these characteristics are far from nightmares. This sharp difference motivates us to take a closer look at the salient features of deep learning, which we single out a few below. 



\subsubsection{Depth}
Deep learning expresses complicated nonlinearity through composing many nonlinear functions; see~(\ref{model:1}). The rationale for this multilayer structure is that, in many real-world datasets such as images, there are different levels of features and lower-level features are building blocks of higher-level ones. See~\cite{yosinski2015understanding} for a visualization of trained features of convolutional neural nets; here in Figure~\ref{fig:vis}, we sample and visualize weights from a pre-trained AlexNet model. This intuition is also supported by empirical results from physiology and neuroscience~\citep{hubel1962receptive, abbasi2018deeptune}. The use of function composition marks a sharp difference from traditional statistical methods such as projection pursuit models \citep{friedman1981projection} and multi-index models \citep{li1991sliced, cook2007fisher}. It is often observed that depth helps efficiently extract features that are representative of a dataset. In comparison, increasing width (e.g., number of basis functions) in a shallow model leads to less improvement. This suggests that deep learning models excel at representing a very different function space that is suitable for complex datasets.

\subsubsection{Algorithmic regularization}
The statistical performance of neural networks (e.g., test accuracy) depends heavily on the particular optimization algorithms used for training~\citep{NIPS2017_7003}. This is very different from many classical statistical problems, where the related optimization problems are less complicated. For instance, when the associated optimization problem has a relatively simple structure (e.g., convex objective functions, linear constraints), the solution to the optimization problem can often be unambiguously computed and analyzed. However, in deep neural networks, due to over-parametrization, there are usually many local minima with different statistical performance \citep{li2018visualizing}. Nevertheless, common practice runs stochastic gradient descent with random initialization and finds model parameters with very good prediction accuracy. 


\subsubsection{Implicit prior learning}
It is well observed that deep neural networks trained with only the raw inputs (e.g., pixels of images) can provide a useful representation of the data. This means that after training, the units of deep neural networks can represent features such as edges, corners, wheels, eyes, etc.; see~\cite{yosinski2015understanding}. Importantly, the training process is automatic in the sense that no human knowledge is involved (other than hyper-parameter tuning). This is very different from traditional methods, where algorithms are designed after structural assumptions are posited. 
It is likely that training an over-parametrized model efficiently learns and incorporates the prior distribution $p(\xx)$ of the input, even though deep learning models are themselves discriminative models. With automatic representation of the prior distribution, deep learning typically performs well on similar datasets (but not very different ones) via transfer learning.

\begin{figure}
\centering
\begin{tabular}{cc}
\includegraphics[width = 0.45\textwidth]{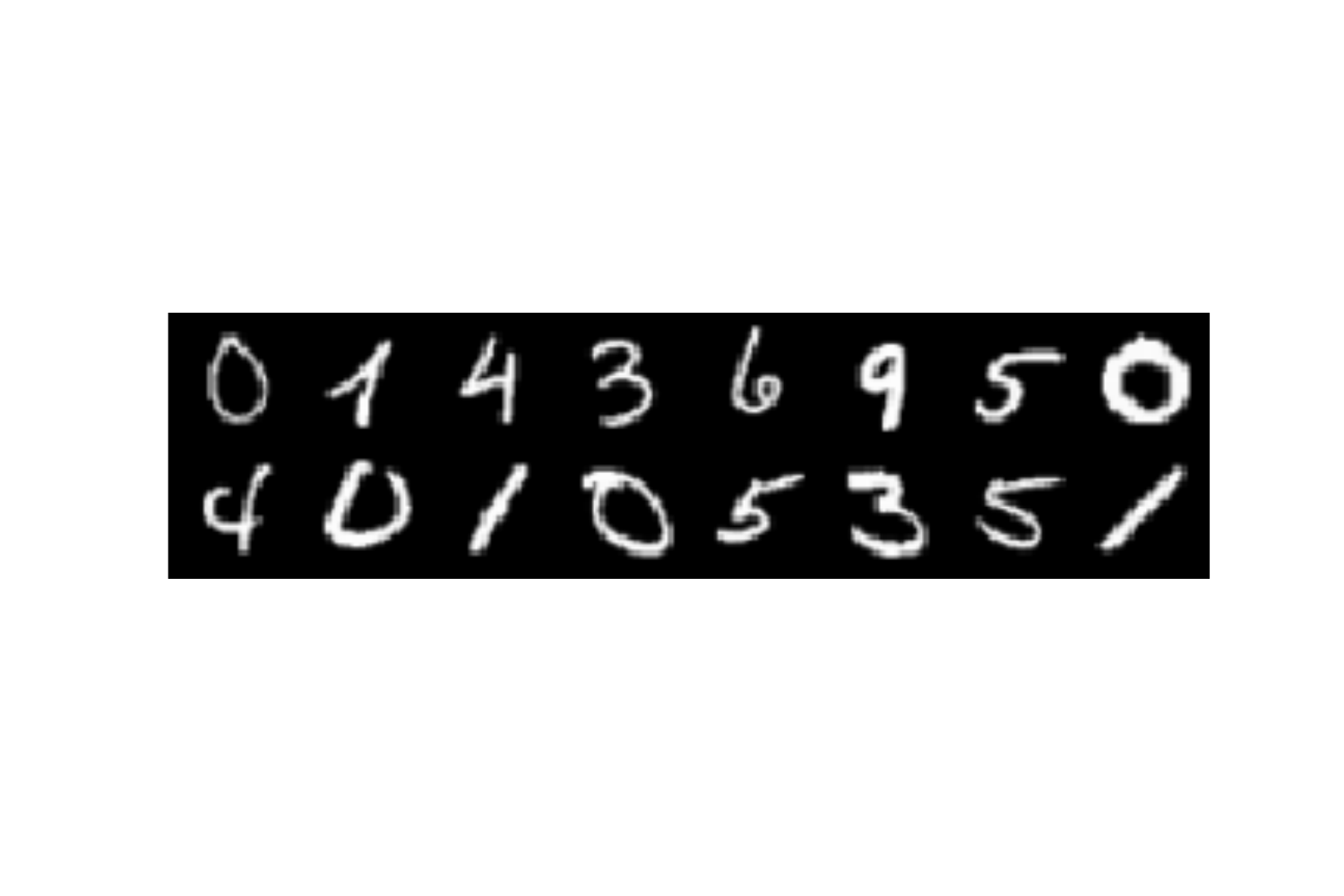} & \includegraphics[width = 0.45\textwidth]{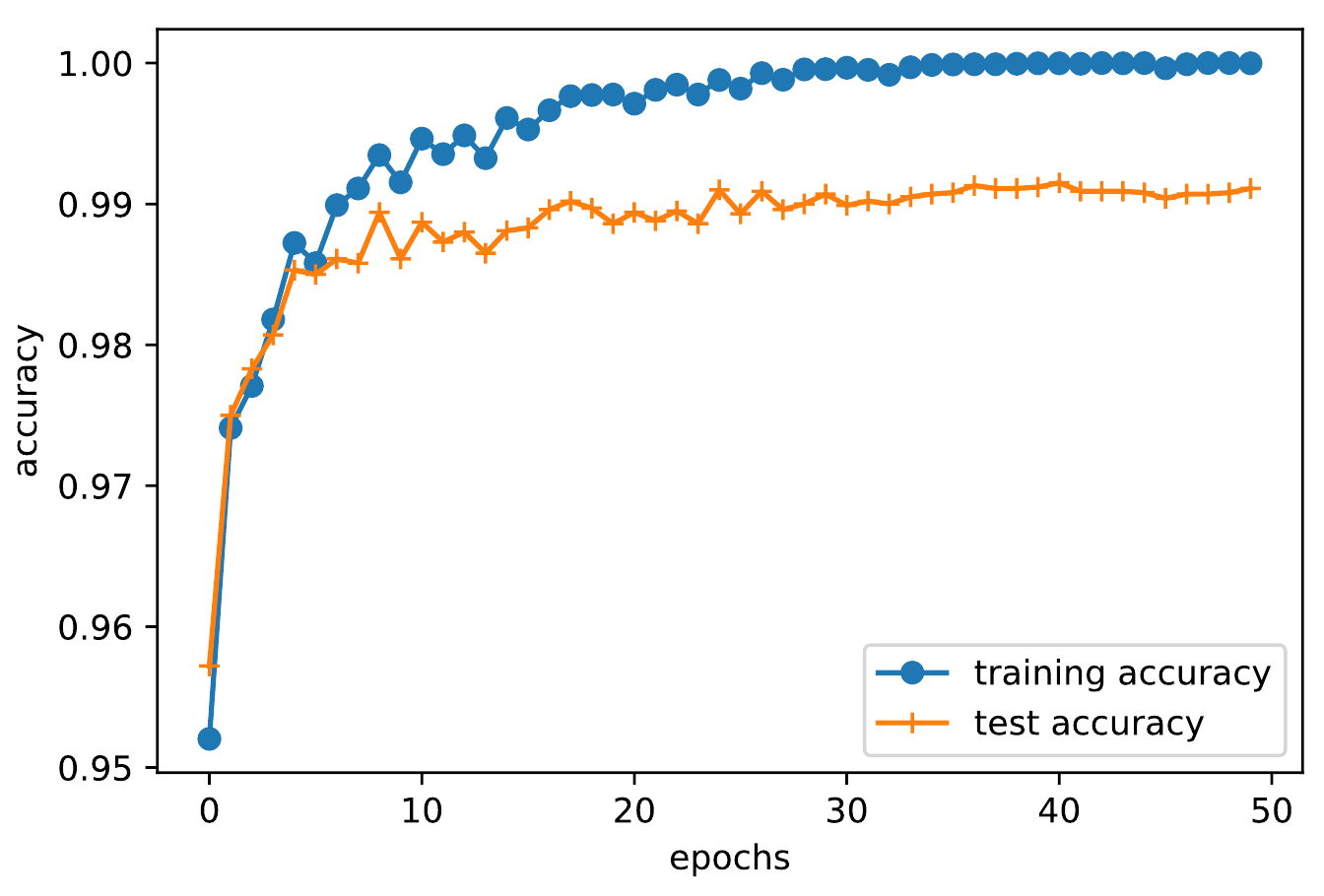}    \tabularnewline
(a) MNIST images & (b) training and test accuracies
\end{tabular}
\caption{(a) shows the images in the public dataset MNIST; and (b) depicts the training and test accuracies along the training dynamics. Note that the training accuracy is approaching $100\%$ and the test accuracy is still high (no overfitting). }\label{fig:mnist}
\end{figure}

\subsection{Towards theory of deep learning}

Despite the empirical success, theoretical support for deep learning is still in its infancy. Setting the stage, for any classifier $f$, denote by $\mathbb{E}(f)$ the expected risk on fresh sample (a.k.a.~test error, prediction error or generalization error), and by $\mathbb{E}_n(f)$ the empirical risk$\,$/$\,$training error averaged over a training dataset. Arguably, the key theoretical question in deep learning is 
\begin{center}
\emph{why is $\mathbb{E}(\hat{f}_{n})$ small, where $\hat{f}_{n}$ is the classifier returned by the training algorithm?}
\end{center}

We follow the conventional approximation-estimation decomposition (sometimes, also bias-variance tradeoff) to decompose the term $\mathbb{E}(\hat{f}_{n})$ into two parts.
Let $\cF$ be the function space expressible by a family of neural nets.
Define $f^* = \argmin_f \mathbb{E}(f)$ to be the best possible classifier and $f^*_{\cF} = \argmin_{f \in \cF} \mathbb{E}(f)$ to be the best classifier in $\cF$. Then, we can decompose the excess error $\cE \triangleq \mathbb{E}(\hat f_n) - \mathbb{E}(f^*)$ into two parts:
\begin{equation}
\cE = \underbrace{\mathbb{E}(f^*_{\cF}) - \mathbb{E}(f^*)}_{\text{approximation error}} ~ + ~ \underbrace{\mathbb{E}(\hat f_n) - \mathbb{E}(f^*_{\cF})}_{\text{estimation error}}.\label{eq:error_decomposition}
\end{equation}
Both errors can be small for deep learning (cf. Figure~\ref{fig:mnist}), which we explain below.
\begin{itemize}
\item{The \emph{approximation error} is determined by the function class $\cF$. Intuitively, the larger the class, the smaller the approximation error. Deep learning models use many layers of nonlinear functions (Figure~\ref{fig:FFNN})that can drive this error small. Indeed, in Section~\ref{sec:approx}, we provide recent theoretical progress of its representation power. For example, deep models allow efficient representation of interactions among variable while shallow models cannot.
}
\item{The \emph{estimation error} reflects the generalization power, which is influenced by both the complexity of the function class $\mathcal{F}$ and the properties of the training algorithms. Interestingly, for \emph{over-parametrized} deep neural nets, stochastic gradient descent typically results in a near-zero  training error (i.e., $\mathbb{E}_{n}(\hat{f}_{n})\approx 0$; see e.g. left panel of Figure~\ref{fig:mnist}). Moreover, its generalization error $\mathbb{E}(\hat{f}_{n})$ remains small or moderate. This ``counterintuitive'' behavior suggests that for over-parametrized models, gradient-based algorithms enjoy benign statistical properties; we shall see in Section~\ref{sec:generalization} that gradient descent enjoys \textit{implicit regularization} in the over-parametrized regime even without explicit regularization (e.g., $\ell_2$ regularization).
}
\end{itemize}

The above two points lead to the following heuristic explanation of the success of deep learning models. The large depth of deep neural nets and heavy over-parametrization lead to small or zero training errors, even when running simple algorithms with moderate number of iterations. In addition, these simple algorithms with moderate number of steps do not explore the entire function space and thus have limited complexities, which results in small generalization error with a large sample size. Thus, by combining the two aspects, it explains heuristically that the test error is also small.

\subsection{Roadmap of the paper}

We first introduce basic deep learning models in Sections~\ref{sec:super}--\ref{sec:unsup}, and then examine their representation power via the lens of approximation theory in Section~\ref{sec:approx}. Section~\ref{sec:opt} is devoted to training algorithms and their ability of driving the training error small. Then we sample recent theoretical progress towards demystifying the generalization power of deep learning in Section~\ref{sec:generalization}. Along the way, we provide our own perspectives, and at the end we identify a few interesting questions for future research in Section~\ref{sec:discuss}. The goal of this paper is to present suggestive methods and results, rather than giving conclusive arguments (which is currently unlikely) or a comprehensive survey. We hope that our discussion serves as a stimulus for new statistics research.

\section{Feed-forward neural networks}\label{sec:super}

Before introducing the vanilla feed-forward neural nets, let us set up necessary notations for the rest of this section. We focus primarily on classification problems, as regression problems can be addressed similarly. Given the training dataset $\{ (y_i, \xx_i )\}_{1\leq i\leq n}$ where $y_i \in [K] \triangleq \{1,2,\ldots,K\}$ and $\xx_{i} \in \R^d$ are independent across $i \in [n]$, supervised learning aims at finding a (possibly random) function $\hat f(\xx)$ that predicts the outcome $y$ for a new input $\xx$, assuming $(y, \xx)$ follows the same distribution as $(y_i, \xx_i )$. In the terminology of machine learning, the input $\xx_i$ is often called the \textit{feature}, the output $y_i$ called the \textit{label}, and the pair $(y_i,\bm{x}_i)$ is an \textit{example}. The function $\hat f$ is called the \textit{classifier}, and estimation of $\hat f$ is \textit{training} or \textit{learning}. The performance of $\hat f$ is evaluated through the prediction error $\P(y \neq \hat f(\xx))$, which can be often estimated from a separate test dataset.

As with classical statistical estimation, for each $k \in [K]$, a classifier approximates the conditional probability $\P(y = k | \xx)$ using a function $f_k(\xx; \btheta_k)$ parametrized by $\btheta_k$. Then the category with the highest probability is predicted. Thus, learning is essentially estimating the parameters $\btheta_k$. In statistics, one of the most popular methods is (multinomial) logistic regression, which stipulates a specific form for the functions $f_k(\xx; \btheta_k)$: let $z_k = \xx^\top \bbeta_k + \alpha_k$ and $f_k(\xx; \btheta_k) = Z^{-1} \exp(z_k)$ where $Z = \sum_{k=1}^K \exp(z_k)$ is a normalization factor to make $\{f_k(\xx; \btheta_k)\}_{1\leq k \leq K}$ a valid probability distribution. It is clear that logistic regression induces linear decision boundaries in $\mathbb{R}^{d}$, and hence it is restrictive in modeling nonlinear dependency between $y$ and $\xx$. The deep neural networks we introduce below provide a flexible framework for modeling nonlinearity in a fairly general way.

\subsection{Model setup}

\begin{figure}
\centering
\includegraphics[scale=0.4]{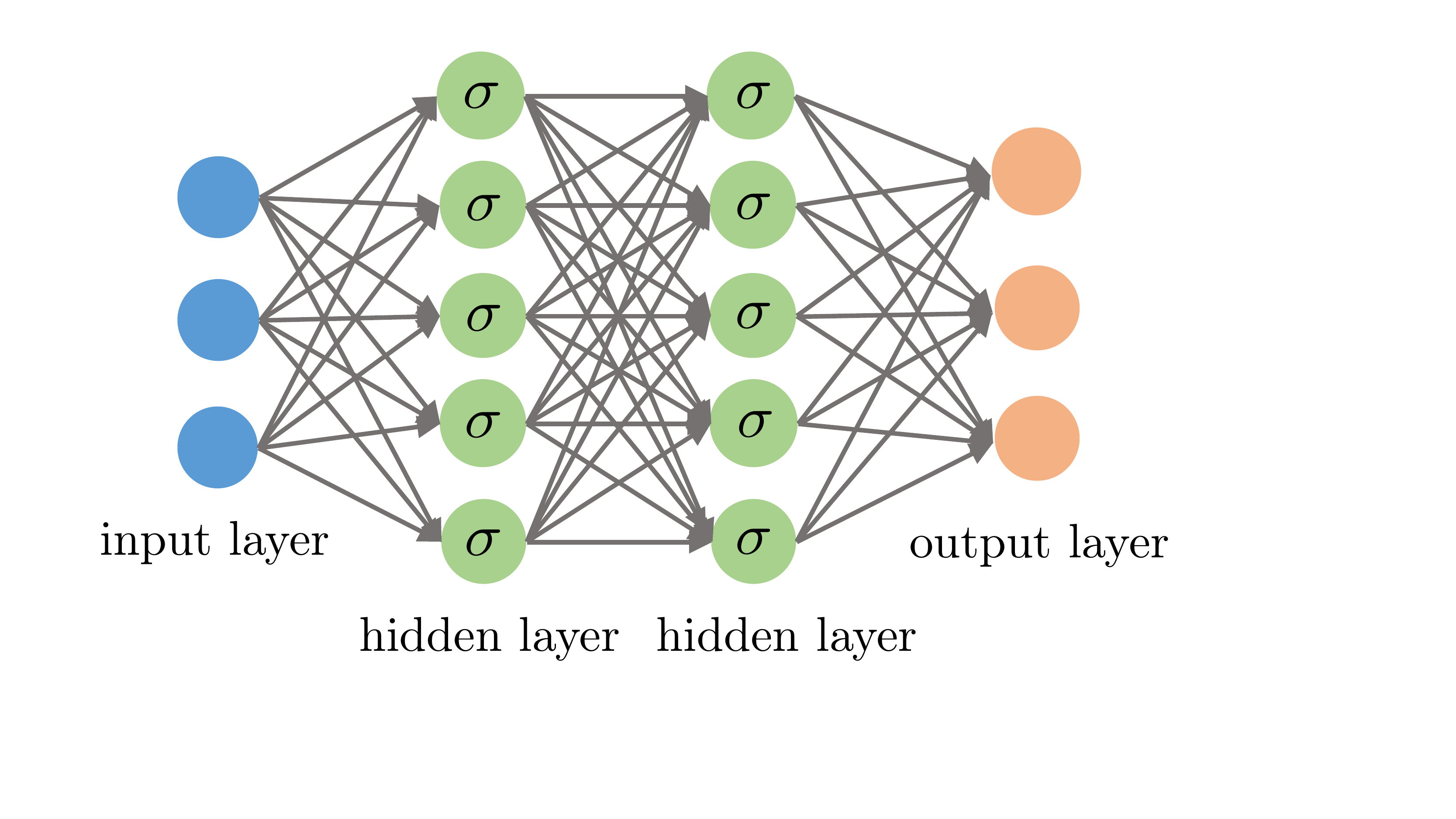}\caption{A feed-forward neural network with an input layer, two hidden layers and an output layer. The input layer represents raw features $\{\bm{x}_{i}\}_{1\leq i\leq n}$. Both hidden layers compute an affine transform (a.k.s. indices) of the input and then apply an element-wise activation function $\bsigma(\cdot)$. Finally, the output returns a linear transform followed by the softmax activation (resp.~simply a linear transform) of the hidden layers for the classification (resp.~regression) problem. \label{fig:FFNN}}
\end{figure}

From the high level, deep neural networks (DNNs) use composition of a series of simple nonlinear functions to model nonlinearity
\begin{equation*}
\hh^{(L)} = \bg^{(L)} \circ  \bg^{(L-1)} \circ \ldots \circ \bg^{(1)} (\xx),
\end{equation*}
where $\circ$ denotes composition of two functions and $L$ is the number of hidden layers, and is usually called \emph{depth} of a NN model. Letting $\bm{h}^{(0)}\triangleq\bm{x}$, one can recursively define
$\hh^{(l)} =  \bg^{(l)} \big(\bm{h}^{(l-1)}\big)$ for all $\ell = 1,2,\ldots, L$. The \textit{feed-forward neural networks}, also called the \textit{multilayer perceptrons} (MLPs), are neural nets with a specific choice of $\bg^{(l)}$: for $\ell = 1,\ldots,L$, define
\begin{equation}\label{eq:fc}
\hh^{(\ell)} = \bg^{(l)} \big(\bm{h}^{(l-1)}\big) \triangleq \bsigma \big(\bW^{(\ell)} \hh^{(\ell-1)} + \bb^{(\ell)}  \big),
\end{equation}
where $\bW^{(l)}$ and  $\bb^{(l)}$ are the weight matrix and the bias$\,$/$\,$intercept, respectively, associated with the $l$-th layer, and $\bsigma(\cdot)$ is usually a simple given (known) nonlinear function called the \textit{activation function}. In words, in each layer $\ell$, the input vector $\hh^{(\ell-1)}$ goes through an affine transformation first and then passes through a fixed nonlinear function $\bsigma(\cdot)$. See Figure~\ref{fig:FFNN} for an illustration of a simple MLP with two hidden layers. The activation function $\bsigma(\cdot)$ is usually applied element-wise, and a popular choice is the ReLU (Rectified Linear Unit) function:
\begin{equation}
[\bsigma(\zz)]_j = \max\{ z_j, 0 \}.
\end{equation}
Other choices of activation functions include leaky ReLU, $\tanh$ function \citep{maas2013rectifier} and the classical sigmoid function $(1+e^{-z})^{-1}$, which is less used now.


Given an output $\hh^{(L)}$ from the final hidden layer and a label $y$, we can define a loss function to minimize. A common loss function for classification problems is the multinomial logistic loss. Using the terminology of deep learning, we say that $\hh^{(L)}$ goes through an affine transformation and then the \textit{soft-max} function:
\begin{equation*}
f_k(\xx; \btheta) \triangleq  \frac{\exp(z_k)}{\sum_k \exp(z_k)}, \quad \forall\, k\in[K], \qquad \where~ \zz = \bW^{(L+1)} \hh^{(L)} + \bb^{(L+1)} \in \mathbb{R}^{K}.
\end{equation*}
Then the loss is defined to be the cross-entropy between the label $y$ (in the form of an indicator vector) and the score vector $ (f_1(\xx; \btheta),\ldots,f_K(\xx; \btheta))^\top$, which is exactly the negative log-likelihood of the multinomial logistic regression model:
\begin{equation}\label{eq:crossentropy}
\mathcal{L}(\ff(\xx; \btheta), y)=-\sum_{k=1}^K \bbone\{y = k\} \log p_k,
\end{equation}
where $\btheta \triangleq \{ \bW^{(\ell)}, \bb^{(\ell)}: 1\leq \ell \leq L+1\}$.
As a final remark, the number of parameters scales with both the depth $L$ and the width (i.e., the dimensionality of $\bW^{(\ell)}$), and hence it can be quite large for deep neural nets. 

\subsection{Back-propagation in computational graphs}

Training neural networks follows the \emph{empirical risk minimization} paradigm that minimizes the loss (e.g., (\ref{eq:crossentropy})) over all the training data. This minimization is usually done via \textit{stochastic gradient descent} (SGD). In a way similar to gradient descent, SGD starts from a certain initial value $\btheta^{0}$ and then iteratively updates the parameters $\btheta^{t}$ by moving it in the direction of the negative gradient.
The difference is that, in each update, a small subsample $\mathcal{B} \subset [n]$ called a \textit{mini-batch}---which is typically of size 32--512---is randomly drawn and the gradient calculation is only on $\mathcal{B}$ instead of the full batch $[n]$.  This saves considerably the computational cost in calculation of gradient.  By the law of large numbers, this stochastic gradient should be close to the full sample one, albeit with some random fluctuations.  A pass of the whole training set is called an \textit{epoch}. Usually, after several or tens of epochs, the error on a validation set levels off and training is complete. See Section~\ref{sec:opt} for more details and variants on training algorithms.

The key to the above training procedure, namely SGD, is the calculation of the gradient $\nabla \ell_{\cB}(\btheta)$, where
\begin{equation}\label{eq:loss-nn}
\ell_{\cB}(\btheta) \triangleq |\cB|^{-1} \sum_{i \in \cB} \mathcal{L}(\ff(\xx_i ; \btheta), y_i).
\end{equation}
Gradient computation, however, is in general nontrivial for complex models, and it is susceptible to numerical instability for a model with large depth. Here, we introduce an efficient approach, namely \textit{back-propagation}, for computing gradients in neural networks.

Back-propagation \citep{rumelhart1985learning} is a direct application of the chain rule in networks. As the name suggests, the calculation is performed in a backward fashion: one first computes $\partial \ell_{\cB}/\partial \hh^{(L)}$, then $\partial \ell_{\cB}/\partial \hh^{(L-1)}$, $\ldots$, and finally $\partial \ell_{\cB}/\partial \hh^{(1)}$. For example, in the case of the ReLU activation function\footnote{The issue of non-differentiability at the origin is often ignored in implementation.}, we have the following recursive$\,$/$\,$backward relation
\begin{equation}\label{eq:grad}
\frac{\partial \ell_{\cB}}{\partial \hh^{(\ell-1)}} =  \frac{\partial \hh^{(\ell)}}{\partial \hh^{(\ell-1)}} \cdot \frac{\partial \ell_{\cB}}{\partial \hh^{(\ell)}} = (\bW^{(\ell)})^\top \mathsf{diag}\left( \bbone\{\bW^{(\ell)} \hh^{(\ell-1)} + \bb^{(\ell)}  \ge \bm{0}\}  \right) \frac{\partial \ell_{\cB}}{\partial \hh^{(\ell)}}
\end{equation}
where $\mathsf{diag}(\cdot)$ denotes a diagonal matrix with elements given by the argument. Note that the calculation of $\partial \ell_{\cB} / \partial \hh^{(\ell-1)}$ depends on $\partial \ell_{\cB} / \partial \hh^{(\ell)}$, which is the partial derivatives from the next layer. In this way, the derivatives are ``back-propagated'' from the last layer to the first layer. These derivatives $\{\partial \ell_{\cB} / \partial \hh^{(\ell)}\}$ are then used to update the parameters. For instance, the gradient update for $\bW^{(\ell)}$ is given by
\begin{equation}\label{eq:Wupdate}
\bW^{(\ell)} \leftarrow \bW^{(\ell)} - \eta \frac{\partial \ell_{\cB}}{\partial \bW^{(\ell)}}, \quad \where\quad \frac{\partial \ell_{\cB}}{\partial W_{jm}^{(\ell)}} = \frac{\partial \ell_{\cB}}{\partial h_j^{(\ell)}} \cdot \sigma' \cdot h_m^{(\ell-1)},
\end{equation}
where $\sigma' = 1$ if the $j$-th element of $\bW^{(\ell)} \hh^{(\ell-1)} + \bb^{(\ell)}$ is nonnegative, and $\sigma' = 0$ otherwise. The step size $\eta >0$, also called the \textit{learning rate}, controls how much parameters are changed in a single update.

\begin{figure}[t]
\centering
\includegraphics[width=0.75\textwidth]{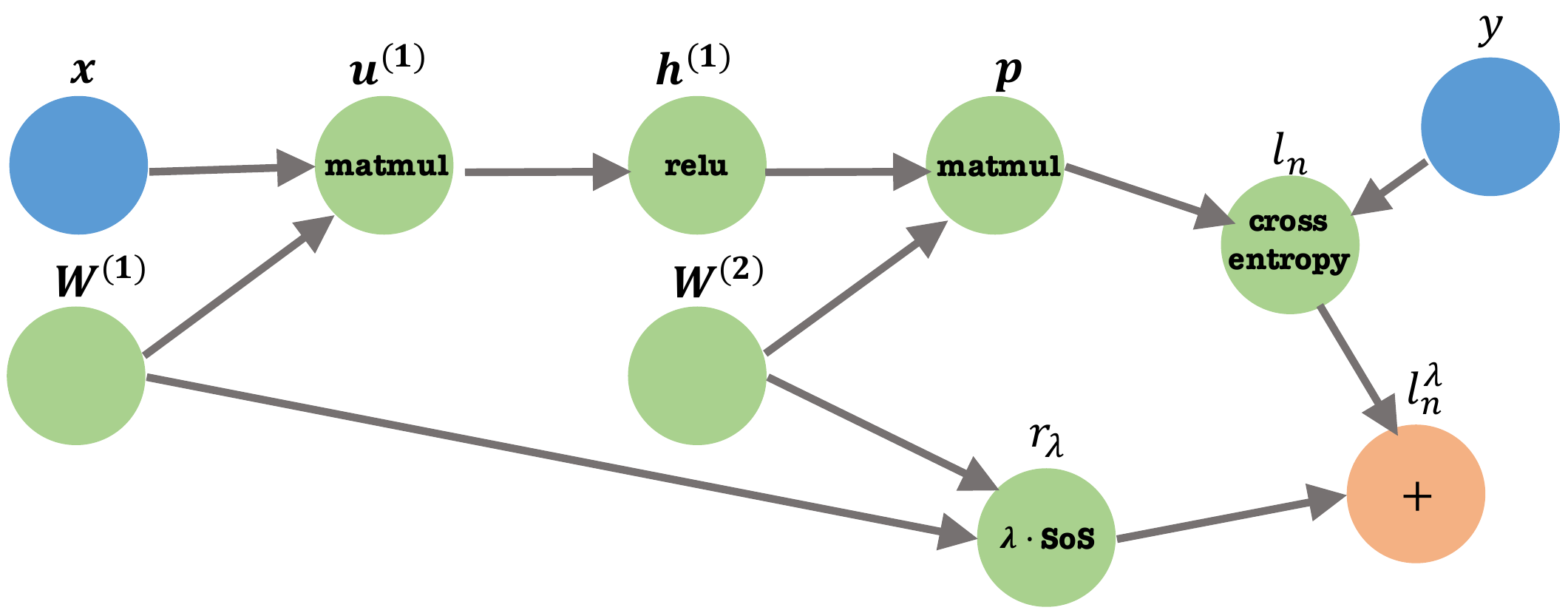}\caption{The computational graph illustrates the loss \eqref{eq:regloss}. For simplicity, we omit the bias terms. Symbols inside nodes represent functions, and symbols outside nodes represent function outputs (vectors/scalars). {\normalfont \texttt{matmul}} is matrix multiplication, {\normalfont \texttt{relu}} is the ReLU activation, {\normalfont \texttt{cross entropy}} is the cross entropy loss, and {\normalfont \texttt{SoS}} is the sum of squares.} \label{fig:comgraph}
\end{figure}

A more general way to think about neural network models and training is to consider \textit{computational graphs}. Computational graphs are directed acyclic graphs that represent functional relations between variables. They are very convenient and flexible to represent function composition, and moreover, they also allow an efficient way of computing gradients. Consider an MLP with a single hidden layer and an $\ell_2$ regularization:
\begin{equation}\label{eq:regloss}
\ell_{\cB}^\lambda (\btheta) = \ell_{\cB}(\btheta) + r_\lambda(\btheta) = \ell_{\cB}(\btheta) + \lambda \Big( \sum_{j,j'} \big(W_{j,j'}^{(1)}\big)^2 + \sum_{j,j'} \big(W_{j,j'}^{(2)}\big)^2 \Big),
\end{equation}
where $\ell_{\cB}(\btheta)$ is the same as \eqref{eq:loss-nn}, and $\lambda \ge 0$ is a tuning parameter. A similar example is considered in~\cite{deeplearningbook}. The corresponding computational graph is shown in Figure~\ref{fig:comgraph}. Each node represents a function (inside a circle), which is associated with an output of that function (outside a circle). For example, we view the term $\ell_{\cB}(\btheta)$ as a result of $4$ compositions: first the input data $\xx$ multiplies the weight matrix $\bW^{(1)}$ resulting in $\uu^{(1)}$, then it goes through the ReLU activation function \texttt{relu} resulting in $\hh^{(1)}$, then it multiplies another weight matrix $\bW^{(2)}$ leading to $\pp$, and finally it produces the cross-entropy with label $y$ as in \eqref{eq:crossentropy}. The regularization term is incorporated in the graph similarly.

A forward pass is complete when all nodes are evaluated starting from the input $\xx$. A backward pass then calculates the gradients of $\ell_{\cB}^\lambda$ with respect to all other nodes in the reverse direction. Due to the chain rule, the gradient calculation for a variable (say, $\partial \ell_{\cB} / \partial \uu^{(1)}$) is simple: it only depends on the gradient value of the variables ($\partial \ell_{\cB} / \partial \hh$) the current node points to, and the function derivative evaluated at the current variable value ($\bsigma'(\uu^{(1)})$). Thus, in each iteration, a computation graph only needs to (1) calculate and store the function evaluations at each node in the forward pass, and then (2) calculate all derivatives in the backward pass.

Back-propagation in computational graphs forms the foundations of popular deep learning programming softwares, including TensorFlow~\citep{tensorflow2015-whitepaper} and PyTorch~\citep{paszke2017automatic}, which allows more efficient building and training of complex neural net models.  

\section{Popular models}\label{sec:pop}

Moving beyond vanilla feed-forward neural networks, we introduce two other popular deep learning models, namely, the convolutional neural networks (CNNs) and the recurrent neural networks
(RNNs). 
One important characteristic shared by the two models is \textit{weight sharing}, that is some model parameters are identical across locations in CNNs or across time in RNNs. This is related to the notion of translational invariance in CNNs and stationarity in RNNs. At the end of this section, we introduce a modular thinking for constructing more flexible neural nets.


\subsection{Convolutional neural networks}\label{sec:CNN}

\begin{figure}
\centering

\includegraphics[width=0.95 \linewidth]{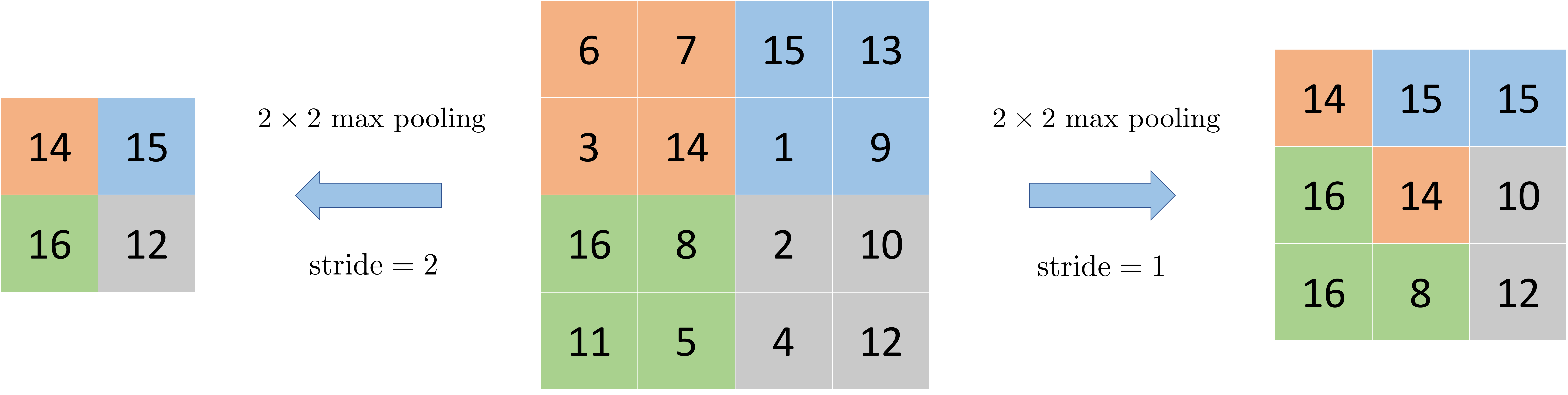}\caption{A $2\times 2$ max pooling layer extracts the maximum of 2 by 2 neighboring pixels$\,$/$\,$features across the spatial dimension. }\label{fig:pooling}
\end{figure}

The convolutional neural network (CNN) \citep{lecun1998gradient, fukushima1982neocognitron} is a special type of feed-forward neural networks that is tailored for image processing. More generally, it is suitable for analyzing data with salient spatial structures. In this subsection, we focus on image classification using CNNs, where the raw input (image pixels) and features of each hidden layer are represented by a 3D tensor $\bm{X}\in\mathbb{R}^{d_{1}\times d_{2}\times d_{3}}$. Here, the first two dimensions $d_1, d_2$ of $\bm{X}$ indicate spatial coordinates of an image while the third $d_3$ indicates the number of channels. For instance, $d_3$ is $3$ for the raw inputs due to the red, green and blue channels, and $d_3$ can be much larger (say, 256) for hidden layers. Each channel is also called a \textit{feature map}, because each feature map is specialized to detect the same feature at different locations of the input, which we will soon explain. 
We now introduce two building blocks of CNNs, namely the convolutional layer and the pooling layer.
\begin{enumerate}

\item \emph{Convolutional layer (CONV)}. A convolutional layer has the same functionality as described in~(\ref{eq:fc}), where the input feature $\bm{X}\in\mathbb{R}^{d_1 \times d_2 \times d_3}$ goes through an affine transformation first and then an element-wise nonlinear activation. The difference lies in the specific form of the affine transformation. A convolutional layer uses a number of \emph{filters} to extract local features from the previous input. More precisely, each filter is represented by a 3D tensor $\bm{F}_{k}\in\mathbb{R}^{w\times w\times d_{3}}$ ($1\leq k\leq \tilde d_3$), where $w$ is the size of the filter (typically 3 or 5) and $\tilde d_3$ denotes the total number of filters. Note that the third dimension $d_3$ of $\bm{F}_{k}$ is equal to that of the input feature $\bm{X}$. For this reason, one usually says that the filter has size $w \times w$, while suppressing the third dimension $d_3$. Each filter $\bm{F}_{k}$ then convolves with the input feature $\bm{X}$ to obtain one single feature map $\bm{O}^{k} \in \mathbb{R}^{(d_1 - w +1) \times (d_1 - w +1)} $, where\footnote{To simplify notation, we omit the bias/intercept term associated with each filter.}
\begin{equation}\label{eq:conv}
O^{k}_{ij}= \big\langle \left[\bm{X}\right]_{ij}, \bm{F}_{k} \big\rangle = \sum_{i'=1}^w \sum_{j'=1}^w \sum_{l=1}^{d_3} [\bm{X}]_{i+i'-1, j+j'-1, l} [\bm{F}_{k}]_{i',j',l}.
\end{equation}
Here $[\bm{X}]_{ij}\in\mathbb{R}^{w\times w\times d_{3}}$ is a small ``patch'' of $\bm{X}$ starting at location $(i,j)$. See Figure \ref{fig:Convolution-operation}
for an illustration of the convolution operation. If we view the 3D tensors $[\bm{X}]_{ij}$ and $\bm{F}_{k}$ as vectors, then each filter essentially computes their inner product with a part of $\bm{X}$ indexed by $i,j$ (which can be also viewed as convolution, as its name suggests). One then pack the resulted feature maps $\{\bm{O}^{k}\}$ into a 3D tensor $\bm{O}$ with size $(d_1 - w +1) \times (d_1 - w +1) \times \tilde d_3$, where
\begin{equation}
[\bm{O}]_{ijk} = [\bm{O}^{k}]_{ij}.
\end{equation}
\begin{figure}
\centering

\includegraphics[height=0.4\textwidth]{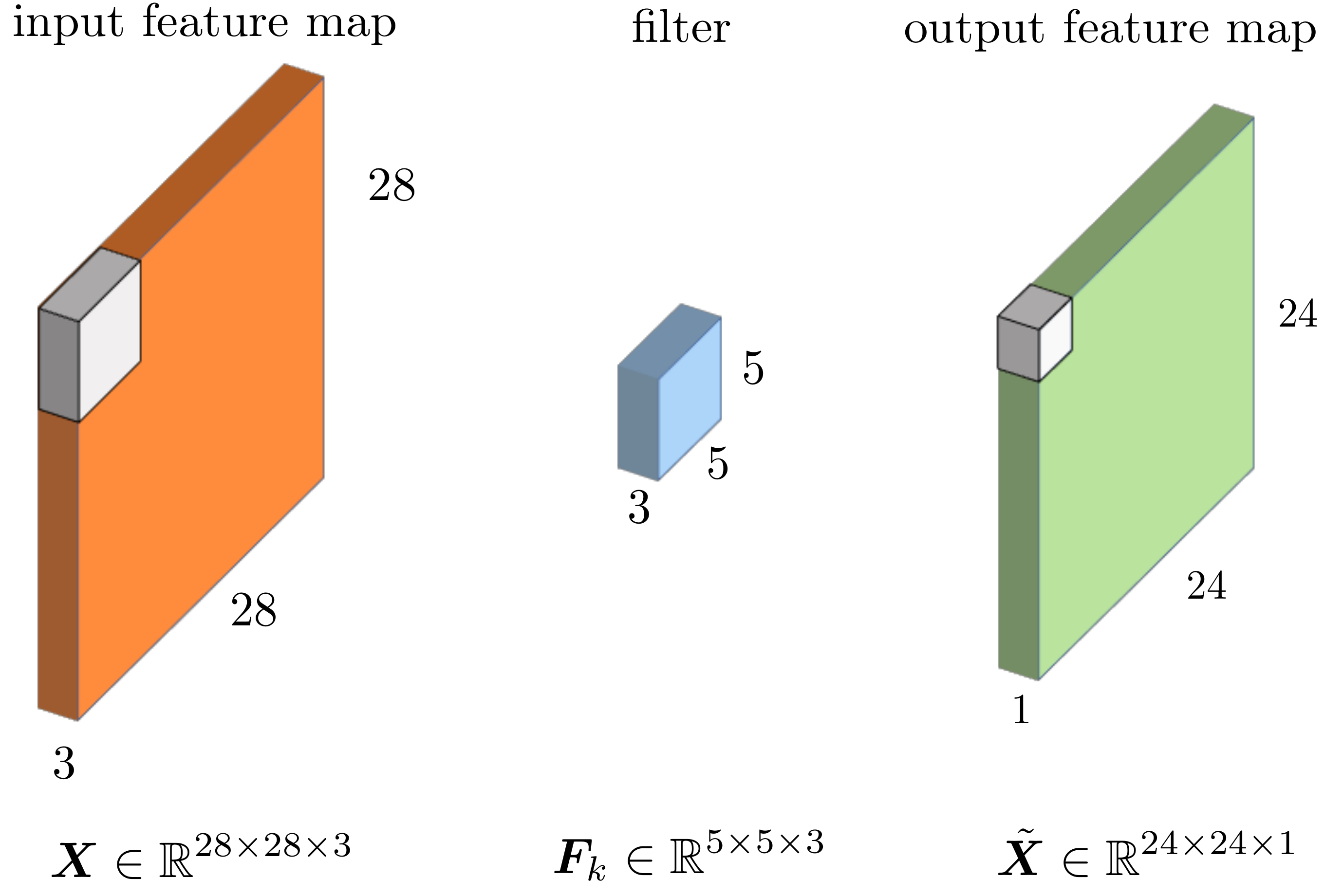}

\caption{$\bm{X}\in \mathbb{R}^{28\times 28 \times 3}$ represents the input feature consisting of $28 \times 28$ spatial coordinates in a total number of 3  channels / feature maps. $\bm{F}_{k}\in\mathbb{R}^{5\times 5 \times 3}$ denotes the $k$-th filter with size $5\times 5$. The third dimension $3$ of the filter automatically matches the number $3$  of channels in the previous input. Every 3D patch of $\bm{X}$ gets convolved with the filter $\bm{F}_{k}$ and this as a whole results in a single output feature map $\tilde{X}_{:,:,k}$ with size $24\times 24\times 1$. Stacking the outputs of all the filters $\{\bm{F}_{k}\}_{1\leq k\leq K}$ will lead to the output feature with size $24\times 24\times K$. \label{fig:Convolution-operation}}

\end{figure}
The outputs of convolutional layers are then followed by nonlinear activation functions. In the ReLU case, we have
\begin{equation}\label{eq:relu}
\tilde{X}_{ijk} = \sigma(O_{ijk}), \qquad \forall\, i \in [d_1-w+1], j \in [d_2-w+1], k \in [\tilde d_3].
\end{equation}
The convolution operation \eqref{eq:conv} and the ReLU activation \eqref{eq:relu} work together to extract features $\tilde{\bm{X}}$ from the input $\bm{X}$. 
Different from feed-forward neural nets, the filters $\bm{F}_k$ are shared across all locations $(i,j)$. A patch $[\bm{X}]_{ij}$ of an input responds strongly (that is, producing a large value) to a filter $\bm{F}_{k}$ if they are positively correlated. Therefore intuitively, each filter $\bm{F}_{k}$ serves to extract features similar to $\bm{F}_{k}$.

As a side note, after the convolution~(\ref{eq:conv}), the spatial size $d_1 \times d_2$ of the input $\bm{X}$ shrinks to ${(d_1-w+1)\times (d_2-w+1)}$ of $\tilde{\bm{X}}$. However one may want the spatial size unchanged. This can be achieved via \emph{padding}, where one appends zeros to the margins of the input $\bm{X}$ to enlarge the spatial size to $(d_1+w-1) \times (d_2+w-1)$. 
In addition, a \emph{stride} in the convolutional layer determines the gap $i' - i$ and $j'-j$ between two patches $\bm{X}_{ij}$ and $\bm{X}_{i'j'}$: in \eqref{eq:conv} the stride is $1$, and a larger stride would lead to feature maps with smaller sizes.

\item \emph{Pooling layer (POOL)}. A pooling layer aggregates the information of nearby features into a single one. This downsampling operation reduces the size of the features for subsequent layers and saves computation. One common form of the pooling layer is composed of the $2 \times 2$ max-pooling filter. It computes $\max \{X_{i,j,k}, X_{i+1,j,k}, X_{i,j+1,k}, X_{i+1,j+1,k} \}$, that is, the maximum of the $2 \times 2$ neighborhood in the spatial coordinates; see Figure \ref{fig:pooling} for an illustration. Note that the pooling operation is done separately for each feature map $k$. As a consequence, a $2 \times 2$ max-pooling filter acting on $\bm{X}\in \mathbb{R}^{d_1 \times d_2 \times d_3}$ will result in an output of size $d_1/2 \times d_2/2 \times d_3$. In addition, the pooling layer does not involve any parameters to optimize. Pooling layers serve to reduce redundancy since a small neighborhood around a location $(i,j)$ in a feature map is likely to contain the same information.
\end{enumerate}

\begin{figure}
\centering
\includegraphics[width=0.9 \textwidth]{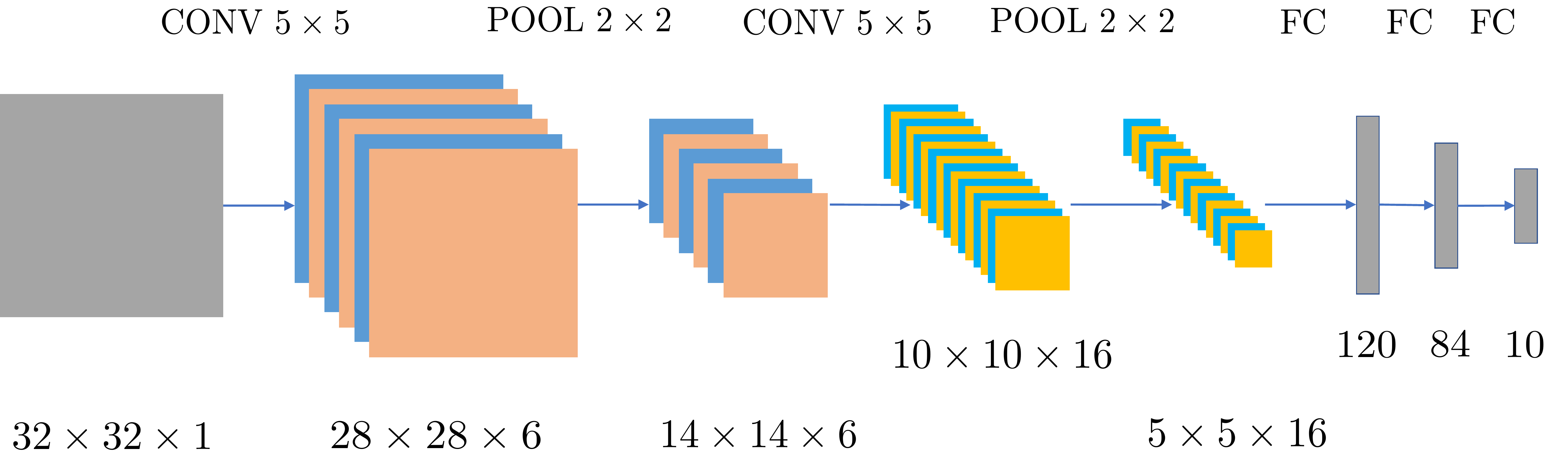}\caption{LeNet is composed of an input layer, two convolutional layers, two pooling layers and three fully-connected layers. Both convolutions are valid and use filters with size $5 \times 5$. In addition, the two pooling layers use $2 \times 2$ average pooling. \label{fig:CNN}}
\end{figure}

In addition, we also use fully-connected layers as building blocks, which we have already seen in Section~\ref{sec:super}. Each fully-connected layer treats input tensor $\bm{X}$ as a vector $\vect(\bm{X})$, and computes $\tilde{\bm{X}} = \bsigma(\bW \vect(\bm{X}))$. A fully-connected layer does not use weight sharing and is often used in the last few layers of a CNN. As an example, Figure~\ref{fig:CNN} depicts the well-known LeNet 5 \citep{lecun1998gradient}, which is composed of two sets of CONV-POOL layers and three fully-connected layers.


%
%
%
%
%
%
%
%

\subsection{Recurrent neural networks}\label{sec:RNN}
Recurrent neural nets (RNNs) are another family of powerful models, which are designed to process time series data and other sequence data. RNNs have successful applications in speech recognition \citep{sak2014long}, machine translation \citep{wu2016google}, genome sequencing \citep{cao2018deep}, etc. The structure of an RNN naturally forms a computational graph, and can be easily combined with other structures such as CNNs to build large computational graph models for complex tasks. Here we introduce vanilla RNNs and improved variants such as long short-term memory (LSTM).


\begin{figure}
\centering
\begin{tabular}{ccc}
\includegraphics[width = 0.3\textwidth]{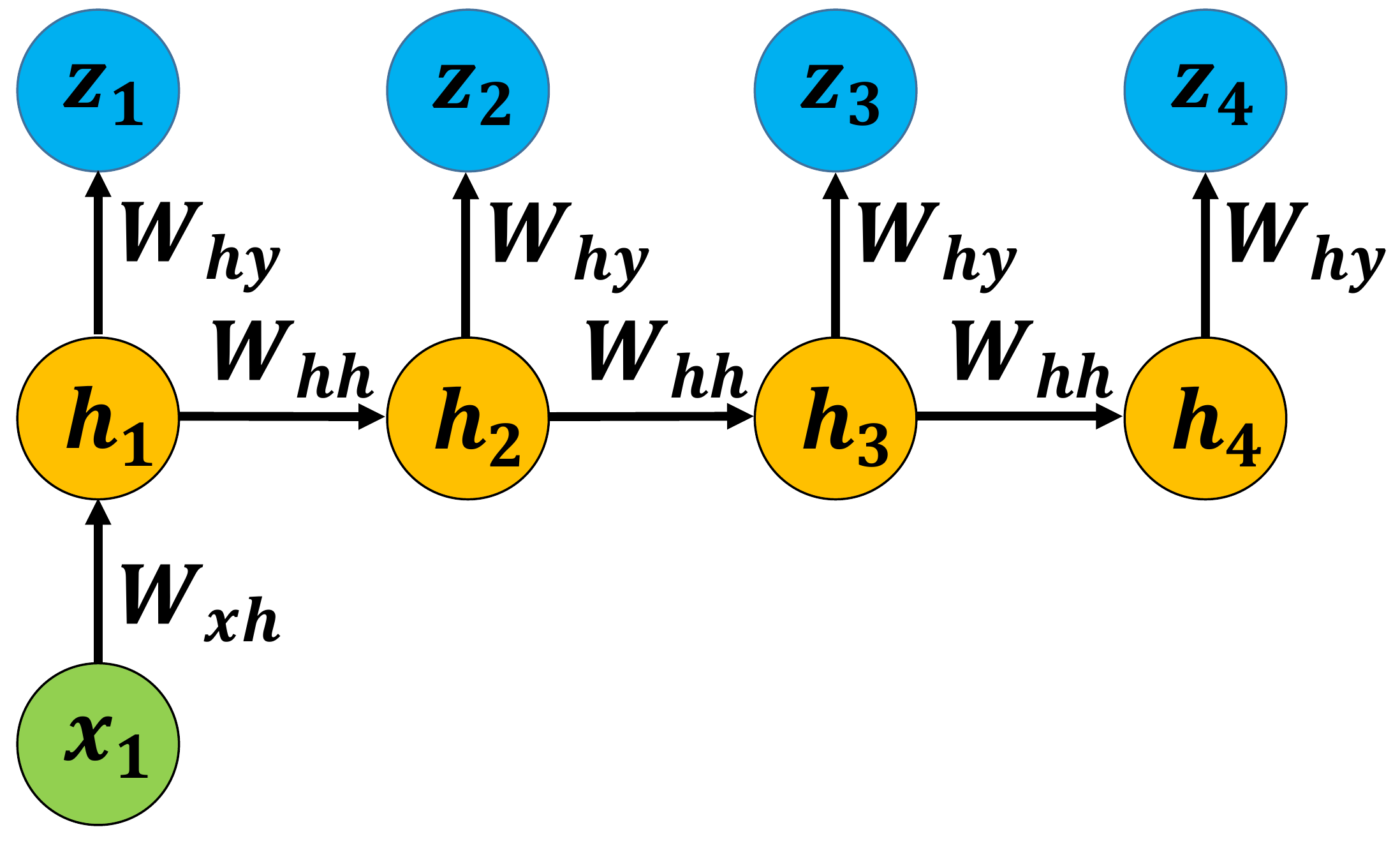} & \includegraphics[width = 0.3\textwidth]{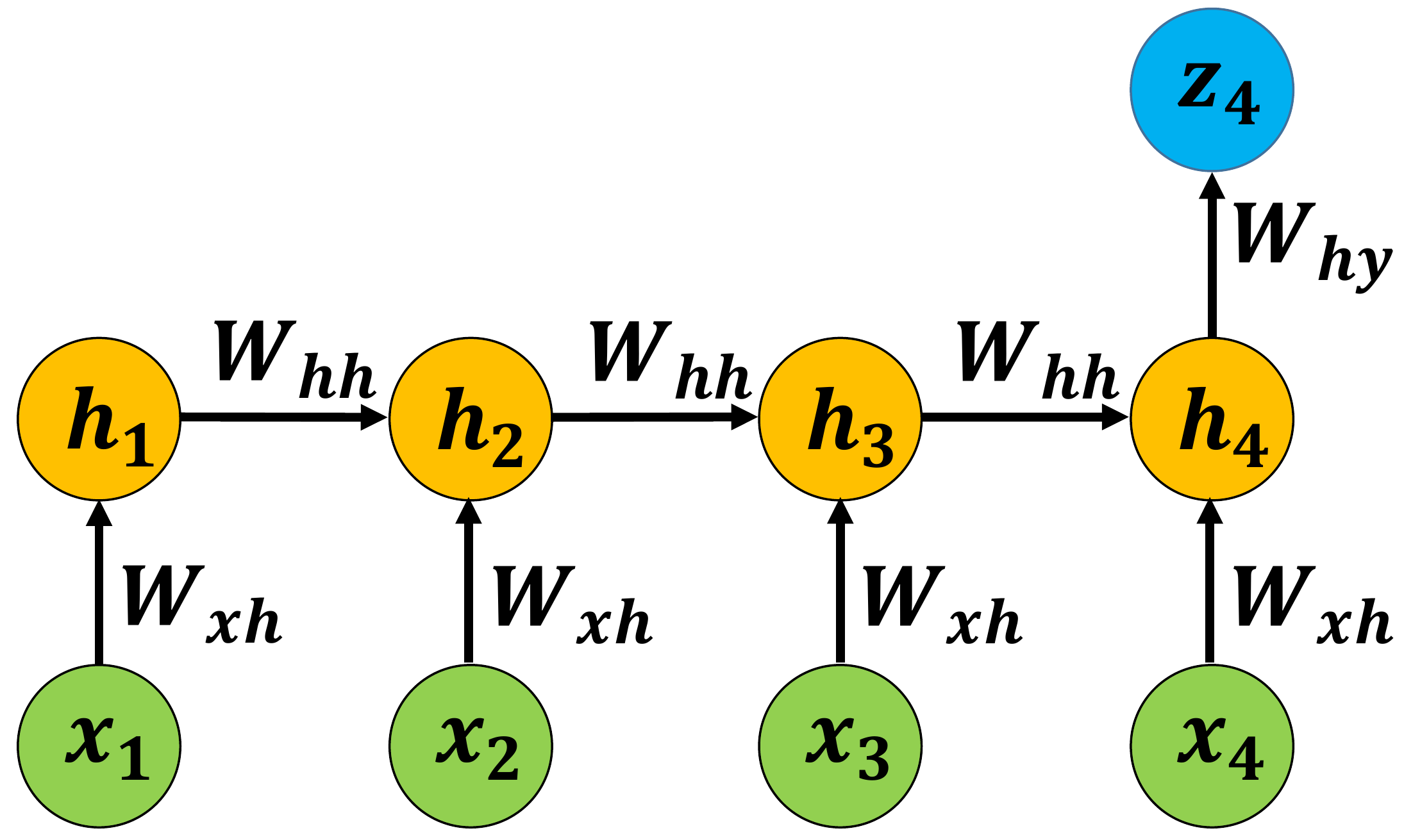} & \includegraphics[width = 0.3\textwidth]{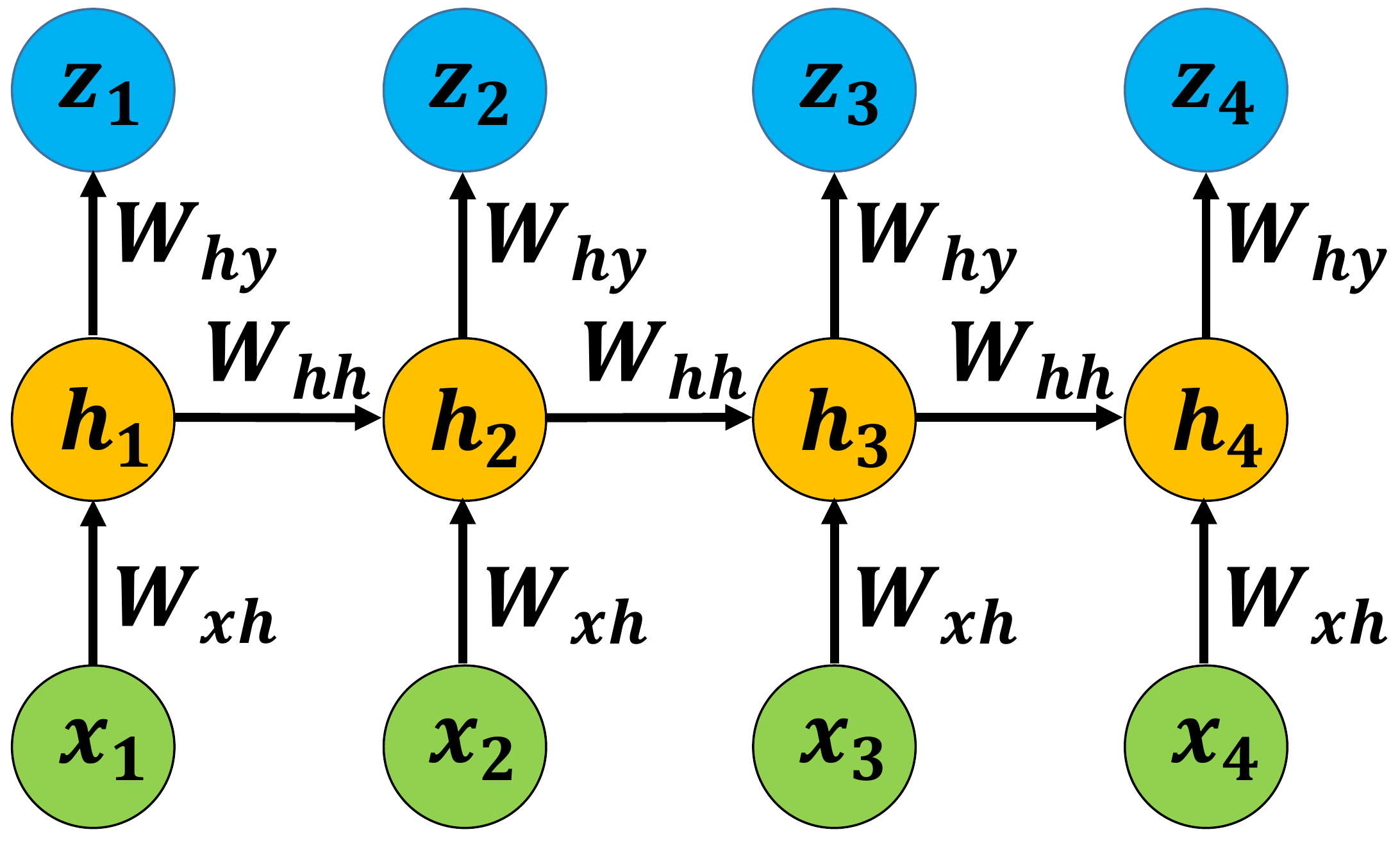} \tabularnewline
(a) One-to-many & (b) Many-to-one & (c) Many-to-many
\end{tabular}
\caption{Vanilla RNNs with different inputs/outputs settings. (a) has one input but multiple outputs; (b) has multiple inputs but one output; (c) has multiple inputs and outputs. Note that the parameters are shared across time steps.}\label{fig:RNN1}
\end{figure}

\subsubsection{Vanilla RNNs}
Suppose we have general time series inputs $\xx_1,\xx_2,\ldots,\xx_T$. A vanilla RNN models the ``hidden state'' at time $t$ by a vector $\hh_t$, which is subject to the recursive formula
\begin{equation}\label{eq:recur}
\hh_t  = \ff_{\btheta}(\hh_{t-1}, \xx_t).
\end{equation}
Here, $f_{\btheta}$ is generally a nonlinear function parametrized by $\btheta$. Concretely, a vanilla RNN with one hidden layer has the following form\footnote{Similar to the activation function $\bsigma(\cdot)$, the function $\btanh(\cdot)$ means element-wise operations.}
\begin{align*}
\hh_t  &= \btanh\left( \bW_{hh} \hh_{t-1}+ \bW_{xh} \xx_t + \bb_\hh \right), \qquad \text{where}~ \tanh(a) = \tfrac{e^{2a} - 1}{e^{2a} + 1}, \\
\zz_t &= \bsigma \left(\bW_{hy} \hh_t + \bb_\zz \right),
\end{align*}
where $\bW_{hh}, \bW_{xh}, \bW_{hy}$ are trainable weight matrices, $\bb_\hh, \bb_\zz$ are trainable bias vectors, and $\zz_t$ is the output at time $t$. Like many classical time series models, those parameters are shared across time. Note that in different applications, we may have different input/output settings (cf.~Figure~\ref{fig:RNN1}). Examples include
\begin{itemize}
\item{ \textbf{One-to-many:} a single input with multiple outputs; see Figure~\ref{fig:RNN1}(a). A typical application is image captioning, where the input is an image and outputs are a series of words.
}
\item{ \textbf{Many-to-one:} multiple inputs with a single output; see Figure~\ref{fig:RNN1}(b). One application is text sentiment classification, where the input is a series of words in a sentence and the output is a label (e.g., positive vs.~negative).
}
\item{ \textbf{Many-to-many:} multiple inputs and outputs; see Figure~\ref{fig:RNN1}(c). This is adopted in machine translation, where inputs are words of a source language (say Chinese) and outputs are words of a target language (say English).
}
\end{itemize}

As the case with feed-forward neural nets, we minimize a loss function using back-propagation, where the loss is typically
\begin{equation*}
\ell_{\cT}(\btheta) = \sum_{t \in \cT} \cL(y_t, \zz_t) = - \sum_{t \in \cT} \sum_{k=1}^K \bbone\{y_t = k\} \log \left( \frac{\exp([\zz_t]_k)}{\sum_k \exp([\zz_t]_k)} \right),
\end{equation*}
where $K$ is the number of categories for classification (e.g., size of the vocabulary in machine translation), and $\cT \subset [T]$ is the length of the output sequence. During the training, the gradients $\partial \ell_{\cT} / \partial \hh_t$ are computed in the reverse time order (from $T$ to $t$). For this reason, the training process is often called \textit{back-propagation through time}. 

One notable drawback of vanilla RNNs is that, they have difficulty in capturing long-range dependencies in sequence data when the length of the sequence is large. This is sometimes due to the phenomenon of \emph{exploding$\,$/$\,$vanishing gradients}. Take Figure~\ref{fig:RNN1}(c) as an example. Computing $\partial \ell_{\cT} / \partial \hh_1$ involves the product $\prod_{t=1}^3 (\partial \hh_{t+1} / \partial \hh_{t})$ by the chain rule. However, if the sequence is long, the product will be the multiplication of many Jacobian matrices, which usually results in exponentially large or small singular values. To alleviate this issue, in practice, the forward pass and backward pass are implemented in a shorter sliding window $\{t_1, t_1+1, \ldots,t_2\}$, instead of the full sequence $\{1,2,\ldots, T\}$. Though effective in some cases, this technique alone does not fully address the issue of long-term dependency. 

\subsubsection{GRUs and LSTM} There are two improved variants that alleviate the above issue: gated recurrent units (GRUs) \citep{cho2014learning} and long short-term memory (LSTM) \citep{hochreiter1997long}.
\begin{itemize}
\item{ A \textbf{GRU} refines the recursive formula \eqref{eq:recur} by introducing \textit{gates}, which are vectors of the same length as $\hh_t$. The gates, which take values in $[0,1]$ elementwise, multiply with $\hh_{t-1}$ elementwise and determine how much they keep the old hidden states.
}
\item{ An \textbf{LSTM} similarly uses gates in the recursive formula. In addition to $\hh_t$, an LSTM maintains a \textit{cell state}, which takes values in $\mathbb{R}$ elementwise and are analogous to counters.
}
\end{itemize}
Here we only discuss LSTM in detail. Denote by $\odot$ the element-wise multiplication. We have a recursive formula in replace of \eqref{eq:recur}:
\begin{align*}
\left( \begin{array}{c} \ii_t \\ \ff_t \\ \oo_t \\ \bgg_t \end{array} \right) &= \left( \begin{array}{c} \bsigma \\ \bsigma \\ \bsigma \\ \btanh \end{array} \right) \bW \left( \begin{array}{c} \hh_{t-1} \\ \xx_t \\ 1 \end{array} \right), \\
\cc_t &= \ff_t \odot \cc_{t-1} + \ii_t \odot \bgg_t, \\
\hh_t &= \oo_t \odot \btanh(\cc_t),
\end{align*}
where $\bW$ is a big weight matrix with appropriate dimensions. The cell state vector $\cc_t$ carries information of the sequence (e.g., singular/plural form in a sentence). The forget gate $\ff_t$ determines how much the values of $\cc_{t-1}$ are kept for time $t$, the input gate $\ii_t$ controls the amount of update to the cell state, and the output gate $\oo_t$ gives how much $\cc_t$ reveals to $\hh_t$.  Ideally, the elements of these gates have nearly binary values. For example, an element of $\ff_t$ being close to $1$ may suggest the presence of a feature in the sequence data. Similar to the skip connections in residual nets, the cell state $\cc_t$ has an additive recursive formula, which helps back-propagation and thus captures long-range dependencies.

\begin{figure}
\centering
\includegraphics[width = 0.4\textwidth]{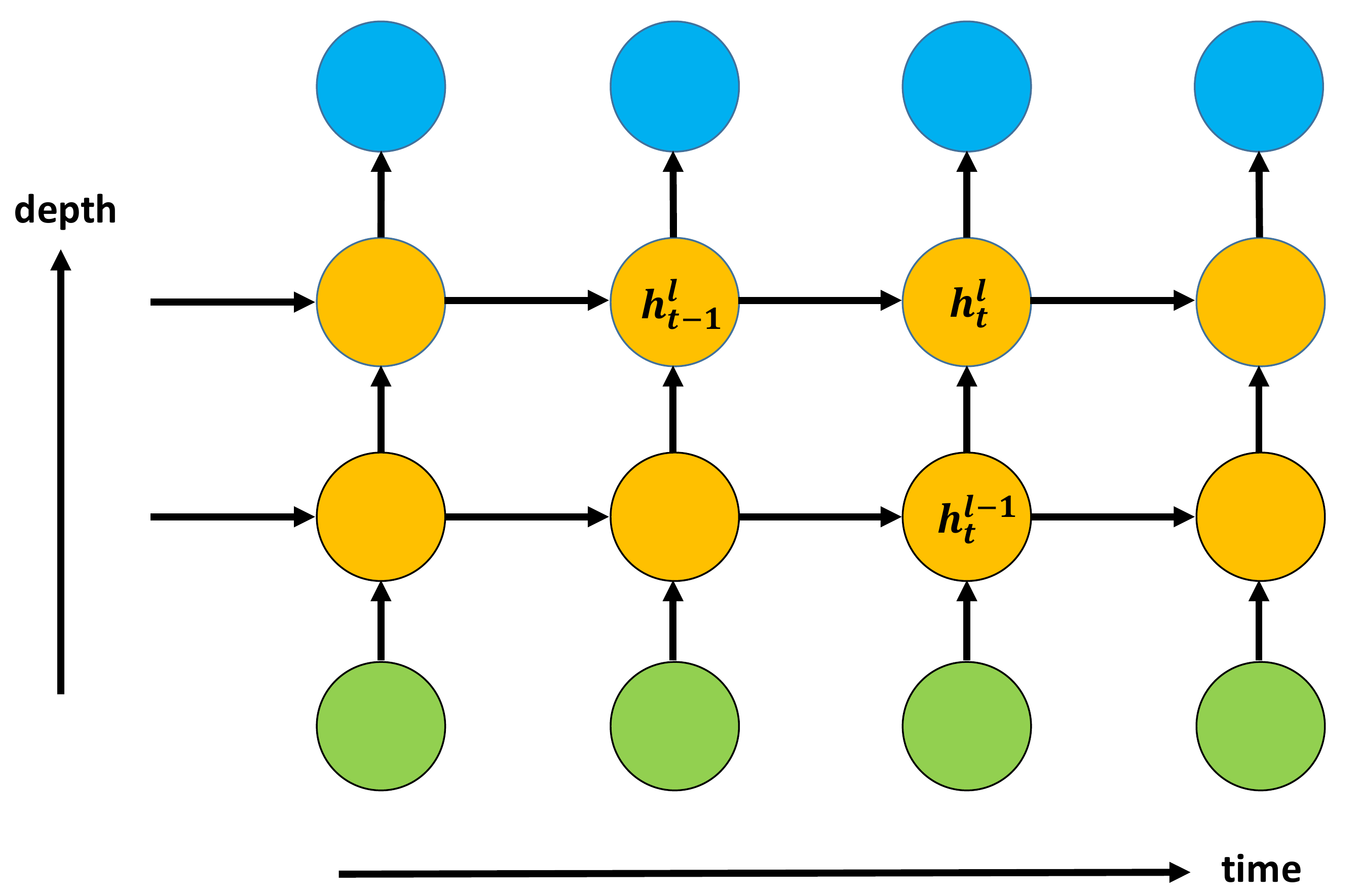}
\caption{A vanilla RNN with two hidden layers. Higher-level hidden states $\hh_t^{\ell}$ are determined by the old states $\hh_{t-1}^\ell$ and lower-level hidden states $\hh_t^{\ell-1}$. Multilayer RNNs generalize both feed-forward neural nets and one-hidden-layer RNNs.}\label{fig:RNN2}
\end{figure}

\subsubsection{Multilayer RNNs} Multilayer RNNs are generalization of the one-hidden-layer RNN discussed above. Figure~\ref{fig:RNN2} shows a vanilla RNN with two hidden layers. In place of \eqref{eq:recur}, the recursive formula for an RNN with $L$ hidden layers now reads
\begin{equation*}
\hh_t^{\ell} =  \btanh \left[\bW^\ell \left( \begin{array}{c} \hh_t^{\ell-1} \\ \hh_{t-1}^\ell \\ 1 \end{array} \right) \right], \quad \text{for all}\, \ell \in [L], \qquad \hh_t^{0} \triangleq \xx_t.
\end{equation*}
Note that a multilayer RNN has two dimensions: the sequence length $T$ and depth $L$. Two special cases are the feed-forward neural nets (where $T=1$) introduced in Section~\ref{sec:super}, and RNNs with one hidden layer (where $L=1$). Multilayer RNNs usually do not have very large depth (e.g., $2$--$5$), since $T$ is already very large.

Finally, we remark that CNNs, RNNs, and other neural nets can be easily combined to tackle tasks that involve different sources of input data. For example, in image captioning, the images are first processed through a CNN, and then the high-level features are fed into an RNN as inputs. Theses neural nets combined together form a large computational graph, so they can be trained using back-propagation. This generic training method provides much flexibility in various applications.

\subsection{Modules}\label{sec:skip}

Deep neural nets are essentially composition of many nonlinear functions. A component function may be designed to have specific properties in a given task, and it can be itself resulted from composing a few simpler functions. In LSTM, we have seen that the building block consists of several intermediate variables, including cell states and forget gates that can capture long-term dependency and alleviate numerical issues. 

This leads to the idea of designing \textit{modules} for building more complex neural net models. Desirable modules usually have low computational costs, alleviate numerical issues in training, and lead to good statistical accuracy. Since modules and the resulting neural net models form computational graphs, training follows the same principle briefly described in Section~\ref{sec:super}.

\begin{figure}[htb!]
\centering
\begin{tabular}{cc}
\includegraphics[scale = 0.5]{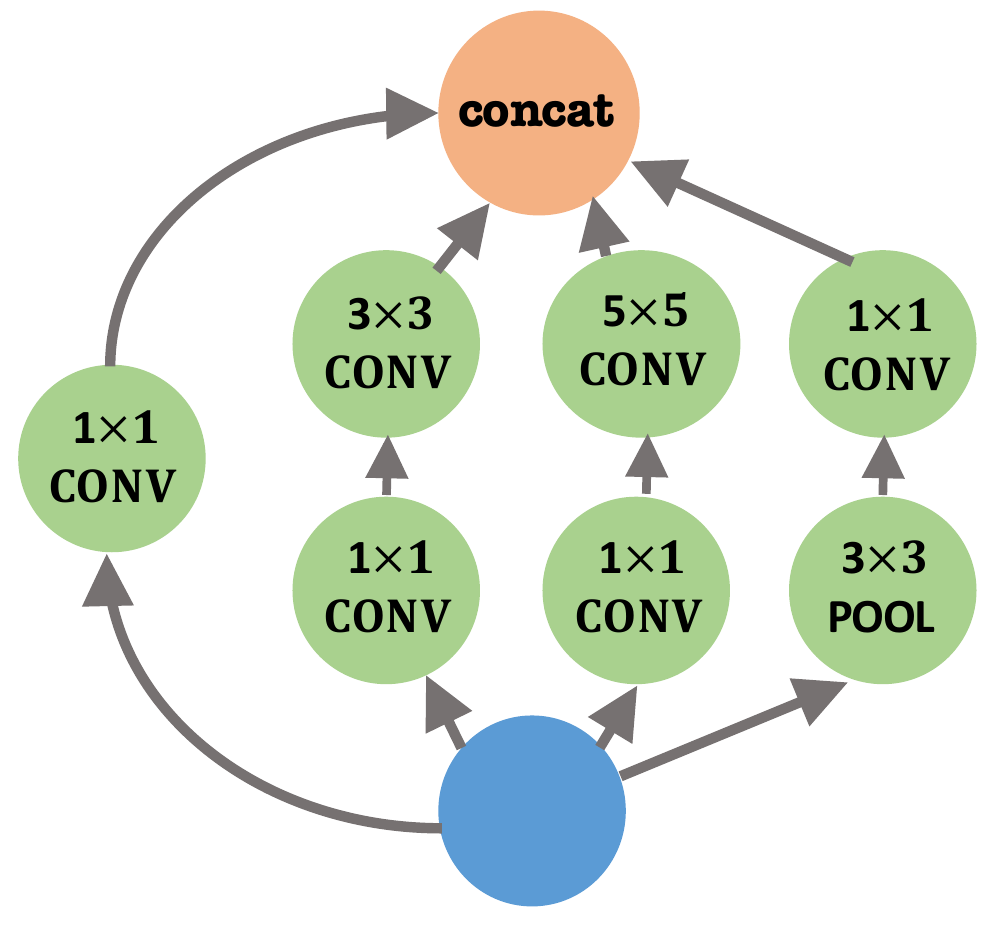} & \includegraphics[scale = 0.5]{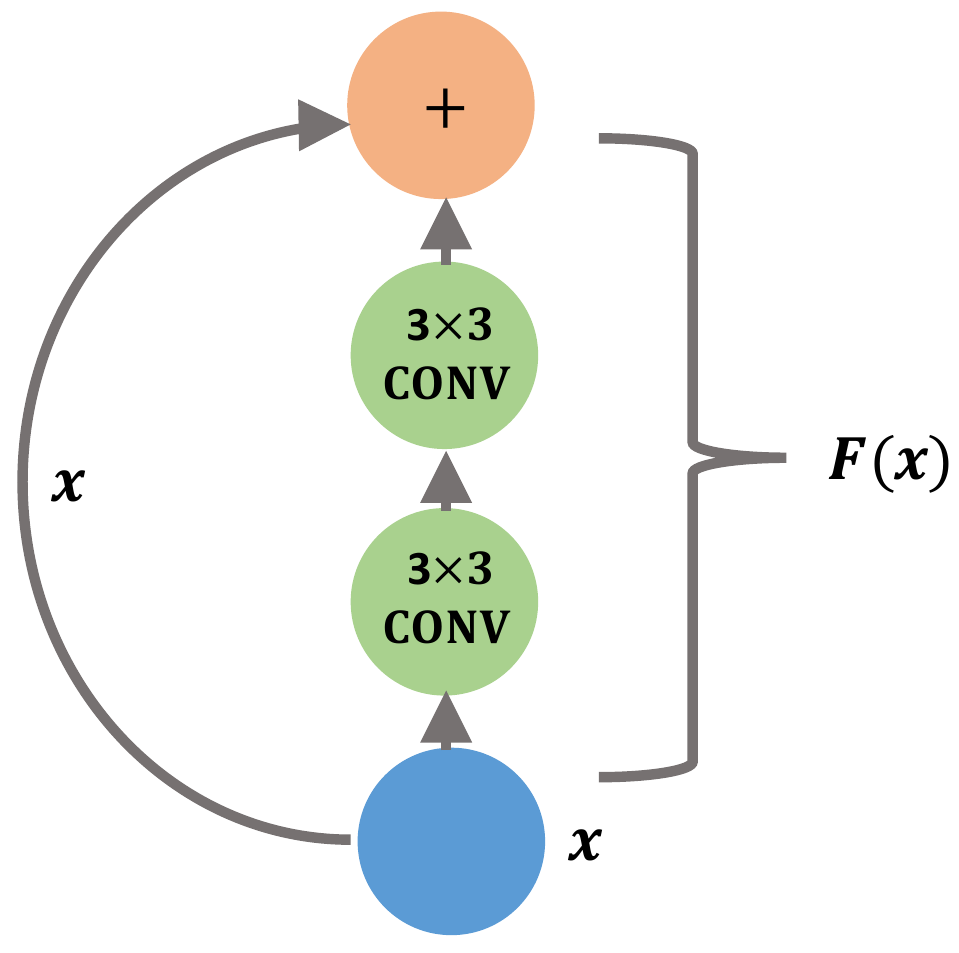} \tabularnewline
(a) ``Inception'' module & (b) Skip connections
\end{tabular}
\caption{(a) The ``Inception'' module from GoogleNet. \texttt{Concat} means combining all features maps into a tensor. (b) Skip connections are added every two layers in ResNets. }\label{fig:skip}
\end{figure}

Here, we use the examples of \emph{Inception} and \emph{skip connections} to illustrate the ideas behind modules. 
Figure~\ref{fig:skip}(a) is an example of ``Inception'' modules used in GoogleNet~\citep{szegedy2015going}. As before, all the convolutional layers are followed by the ReLU activation function. The concatenation of information from filters with different sizes give the model great flexibility to capture spatial information. Note that $1 \times 1$ filters is an $1 \times 1 \times d_3$ tensor (where $d_3$ is the number of feature maps), so its convolutional operation does not interact with other spatial coordinates, only serving to aggregate information from different feature maps at the same coordinate. This reduces the number of parameters and speeds up the computation. Similar ideas appear in other work \citep{lin2013network, iandola2016squeezenet}.


Another module, usually called \textit{skip connections}, is widely used to alleviate numerical issues in very deep neural nets, with additional benefits in optimization efficiency and statistical accuracy. Training very deep neural nets are generally more difficult, but the introduction of skip connections in \emph{residual networks} \citep{he2016deep, he2016identity} has greatly eased the task. 

The high level idea of skip connections is to add an identity map to an existing nonlinear function. Let $\bF(\xx)$ be an arbitrary nonlinear function represented by a (fragment of) neural net, then the idea of skip connections is simply replacing $\bF(\xx)$ with $\xx + \bF(\xx)$. Figure~\ref{fig:skip}(b) shows a well-known structure from residual networks \citep{he2016deep}---for every two layers, an identity map is added:
\begin{equation}\label{eq:mapsto}
\xx \longmapsto \bsigma(\xx + \bF(\xx)) = \bsigma(\xx + \bW' \bsigma(\bW \xx + \bb) + \bb'),
\end{equation}
where $\xx$ can be hidden nodes from any layer and $\bW, \bW', \bb, \bb'$ are corresponding parameters. By repeating (namely composing) this structure throughout all layers, \cite{he2016deep, he2016identity} are able to train neural nets with hundreds of layers easily, which overcomes well-observed training difficulties in deep neural nets. Moreover, deep residual networks also improve statistical accuracy, as the classification error on ImageNet challenge was reduced by $46\%$ from 2014 to 2015. As a side note, skip connections can be used flexibly. 
They are not restricted to the form in \eqref{eq:mapsto}, and can be used between any pair of layers $\ell, \ell'$ \citep{Huang17}. 


\section{Deep unsupervised learning}\label{sec:unsup}

In supervised learning, given labelled training set $\{(y_i,\bx_i)\}$, we focus on discriminative models, which essentially represents $\P(y\,|\,\xx)$ by a deep neural net $f(\xx; \btheta)$ with parameters $\btheta$. Unsupervised learning, in contrast, aims at extracting \emph{information} from \emph{unlabeled} data $\{\bm{x}_{i}\}$, where the labels $\{y_i\}$ are absent. In regard to this information, it can be a low-dimensional embedding of the data $\{ \xx_i \}$ or a generative model with latent variables to approximate the distribution $\P_{\bX}(\xx)$. To achieve these goals, we introduce two popular unsupervised deep leaning models, namely, autoencoders and generative
adversarial networks~(GANs). The first one can be viewed as a dimension reduction technique, and the second as a density estimation method. DNNs are the key elements for both of these two models. 

\subsection{Autoencoders}
Recall that in dimension reduction, the goal is to reduce the dimensionality of the data and at the same time preserve its salient features. In particular, in principal component analysis (PCA), the goal is to embed the data $\{\bm{x}_{i}\}_{1\leq i\leq n}$ into a low-dimensional space via a linear function $\ff$ such that maximum variance can be explained. Equivalently, we want to find linear functions $\ff: \R^d \to \R^k$ and $\bgg: \R^k \to \R^d$ ($k \le d$) such that the difference between $\xx_i$ and $\bgg(\ff(\xx_i))$ is minimized. Formally, we let
\[
\ff\left(\bm{x}\right)=\bm{W}_f\bm{x}\triangleq \hh \quad\text{and}\quad \bgg\left(\bm{h}\right)=\bm{W}_g\bm{h}, \quad \text{where}\quad\bm{W}_f\in\mathbb{R}^{k\times d}\text{ and } \bm{W}_g\in\mathbb{R}^{d\times k}.
\]
Here, for simplicity, we assume that the intercept/bias terms for $\ff$ and $\bgg$ are zero. Then, PCA amounts to minimizing the quadratic loss function
\begin{equation}
\text{minimize}_{\bm{W}_f, \bm{W}_g}\qquad \frac{1}{n} \sum_{i=1}^{n}\left\Vert \bm{x}_i-\bm{W}_f\bm{W}_g\bm{x}_i\right\Vert _{2}^{2}.\label{eq:linear-AE}
\end{equation}
It is the same as minimizing $\| \bX - \bW \bX \|_{\mathrm{F}}^2$ subject to $\rank(\bW) \le k$, where $\bX\in \mathbb{R}^{p\times n}$ is the design matrix. The solution is given by the singular value decomposition of $\bX$ \citep[Thm.~2.4.8]{golub2013matrix}, which is exactly what PCA does. It turns out that PCA is a special case of autoencoders, which is often known as the \textit{undercomplete linear autoencoder}.

\begin{figure}
\centering\includegraphics[scale=0.3]{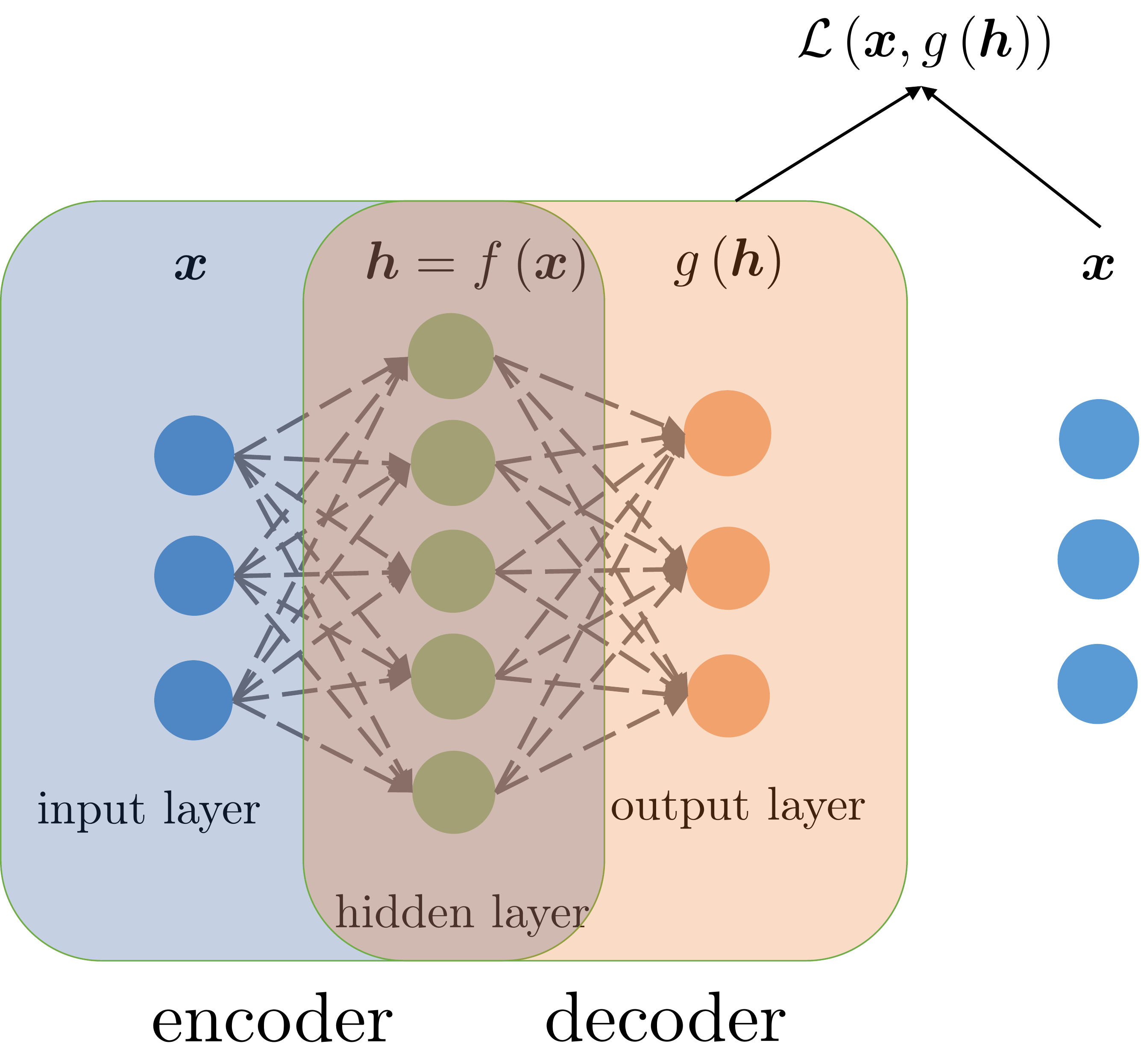}
\caption{First an input $\bm{x}$ goes through the decoder $\ff(\cdot)$, and we obtain its hidden representation $\bm{h}= \ff(\bm{x})$. Then, we use the decoder $\bgg(\cdot)$ to get $\bgg(\bm{h})$ as a reconstruction of $\bm{x}$. Finally, the loss is determined from the difference between the original input $\bm{x}$ and its reconstruction $\bgg(\ff(\bm{x}))$.}\label{fig:AE} 
\end{figure}

More broadly, autoencoders are neural network models for (nonlinear) dimension reduction, which generalize PCA. An autoencoder has two key components, namely, the encoder function $\ff(\cdot)$, which maps the input $\bm{x}\in\mathbb{R}^{d}$ to a hidden code/representation $\bm{h}\triangleq\ff(\bm{x})\in\mathbb{R}^{k}$, and the decoder function $\bgg(\cdot)$, which maps the hidden representation $\bm{h}$ to a point $\bgg(\bm{h})\in\mathbb{R}^{d}$. Both functions can be multilayer neural networks as \eqref{eq:fc}. See Figure~\ref{fig:AE} for an illustration of autoencoders. Let $\mathcal{L}(\bm{x}_{1},\bm{x}_{2})$ be a loss function that measures the difference between $\bm{x}_{1}$ and $\bm{x}_{2}$ in $\R^d$. Similar to PCA, an autoencoder is used to find the encoder $\ff$ and decoder $\bgg$ such that $\mathcal{L}(\bm{x},\bgg(\ff(\bm{x})))$
is as small as possible. Mathematically, this amounts to solving the following minimization problem
\begin{equation}
\mbox{minimize}_{\ff,\bgg} \quad\frac{1}{n}\sum_{i=1}^{n}\mathcal{L}\left(\bm{x}_{i},\bgg\left(\bm{h}_{i}\right)\right) \quad \text{with}\quad\bm{h}_{i}=\ff\left(\bm{x}_{i}\right), \quad \text{for all }\, i \in [n].  \label{eq:AE}
\end{equation}

One needs to make structural assumptions on the functions $\ff$ and $\bg$ in order to find useful representations of the data, which leads to different types of autoencoders. Indeed, if no assumption is made, choosing $\ff$ and $\bgg$ to be identity functions clearly minimizes the above optimization problem. To avoid this trivial solution, one natural way is to require that the encoder $f$ maps the data onto a space with a smaller dimension, i.e., $k < d$. This is the \textit{undercomplete autoencoder} that includes PCA as a special case. There are other structured autoencoders which add desired properties to the model such as sparsity or robustness, mainly through regularization terms. Below we present two other common types of autoencoders.

\begin{itemize}
\item \emph{Sparse autoencoders. } One may believe that the dimension $k$ of the hidden code $\hh_i$ is larger than the input dimension $d$, and that $\hh_i$ admits a sparse representation. As with LASSO \citep{tibshirani1996regression} or SCAD \citep{fan2001variable}, one may add a regularization term to the reconstruction loss $\mathcal{L}$ in \eqref{eq:AE} to encourage sparsity \citep{poultney2007efficient}. A sparse autoencoder solves
\begin{equation*}
\mbox{min}_{\ff,\bgg}  \; \underbrace{ \frac{1}{n}\sum_{i=1}^{n}\mathcal{L}\left(\bm{x}_{i},\bgg\left(\bm{h}_{i}\right)\right) }_{\text{loss}}+ \underbrace{\vphantom{\frac{1}{n}\sum_{i=1}^{n}\mathcal{L}\left(\bm{x}_{i},\bgg\left(\bm{h}_{i}\right)\right)} \lambda\left\Vert \bm{h}_{i}\right\Vert _{1} }_{\text{regularizer}} \quad \text{with} \quad \bm{h}_{i}=\ff\left(\bm{x}_{i}\right), \text{ for all } i \in [n].
\end{equation*}
This is similar to \textit{dictionary learning}, where one aims at finding a sparse representation of input data on an overcomplete basis. Due to the imposed sparsity, the model can potentially learn useful features of the data.

\item \emph{Denoising autoencoders. } One may hope that the model is robust to noise in the data: even if the input data $\bm{x}_i$ are corrupted by small noise $\bxi_i$ or miss some components (the noise level or the missing probability is typically small), an ideal autoencoder should faithfully recover the original data. A denoising autoencoder \citep{vincent2008extracting} achieves this robustness by explicitly building a noisy data $\tilde{\bm{x}}_{i} = \bm{x}_i + \bxi_i$ as the new input, and then solves an optimization problem similar to \eqref{eq:AE} where $\mathcal{L}\left(\bm{x}_{i},\bgg\left(\bm{h}_{i}\right)\right)$ is replaced by $\mathcal{L}\left(\bm{x}_{i},\bgg\left(\ff(\tilde{\bm{x}}_{i})\right)\right)$. A denoising autoencoder encourages the encoder/decoder to be stable in the neighborhood of an input, which is generally a good statistical property. An alternative way could be constraining $f$ and $g$ in the optimization problem, but that would be very difficult to optimize. Instead, sampling by adding small perturbations in the input provides a simple implementation. We shall see similar ideas in Section~\ref{sec:aug}.


\end{itemize}

\subsection{Generative adversarial networks}

Given unlabeled data $\{\bm{x}_{i}\}_{1\leq i\leq n}$, density estimation
aims to estimate the underlying probability density function $\mathbb{P}_{\bm{X}}$
from which the data is generated. Both parametric and nonparametric
estimators \citep{silverman2018density} have been proposed and studied under various assumptions
on the underlying distribution. Different from these classical density estimators, where the density function is explicitly defined in relatively low dimension, generative adversarial networks (GANs) \citep{goodfellow2014generative} can be categorized as an \emph{implicit} density estimator in much higher dimension. The reasons are twofold: (1) GANs put more emphasis on sampling from
the distribution $\mathbb{P}_{\bm{X}}$ than estimation; (2) GANs define the density estimation implicitly through a source distribution $\mathbb{P}_{\bm{Z}}$ and a generator function $g(\cdot)$, which is usually a deep neural network. We introduce GANs from the perspective of sampling from $\mathbb{P}_{\bm{X}}$ and later we will generalize the vanilla GANs using its relation to density estimators.

\subsubsection{Sampling view of GANs}
Suppose the data $\{\bm{x}_{i}\}_{1\leq i\leq n}$ at hand are all real images, and we want to generate \emph{new} natural images.
With this goal in mind, GAN models a \emph{zero-sum} game between two players, namely,
the generator $\mathcal{G}$ and the discriminator $\mathcal{D}$. The
generator $\mathcal{G}$ tries to generate fake images akin to the
true images $\{\bm{x}_{i}\}_{1\leq i\leq n}$ while the discriminator
$\mathcal{D}$ aims at differentiating the fake ones from
the true ones. Intuitively, one hopes to learn a generator $\mathcal{G}$ to generate images where the \emph{best} discriminator $\mathcal{D}$ cannot distinguish. Therefore the payoff is higher for the generator~$\mathcal{G}$ if the probability of the discriminator $\mathcal{D}$
getting wrong is higher, and correspondingly the payoff for the discriminator
correlates positively with its ability to tell wrong from truth.

Mathematically, the generator $\mathcal{G}$ consists of two components,
an source distribution $\mathbb{P}_{\bm{Z}}$ (usually a standard multivariate Gaussian distribution with hundreds of dimensions) and a function $\bg(\cdot)$ which maps a sample
$\bm{z}$ from $\mathbb{P}_{\bm{Z}}$ to a point $\bg(\bm{z})$ living
in the same space as $\bm{x}$. For generating images, $\bg(\bm{z})$ would be a 3D tensor. Here $\bg(\bm{z})$ is the fake sample
generated from $\mathcal{G}$. Similarly the discriminator $\mathcal{D}$
is composed of one function which takes an image ${\bm{x}}$ (real or fake)
and return a number $d({\bm{x}})\in[0,1]$, the probability
of ${\bm{x}}$ being a real sample from $\mathbb{P}_{\bm{X}}$ or not.
Oftentimes, both the generating function $\bg(\cdot)$ and the discriminating
function $d(\cdot)$ are realized by deep neural networks, e.g., CNNs introduced in Section~\ref{sec:CNN}. See Figure~\ref{fig:GAN} for an illustration
for GANs. Denote $\btheta_{\mathcal{G}}$ and $\btheta_{\mathcal{D}}$
the parameters in $\bg(\cdot)$ and $d(\cdot)$, respectively. Then
GAN tries to solve the following \emph{min-max }problem:
\begin{figure}
\centering\includegraphics[width=0.9\textwidth]{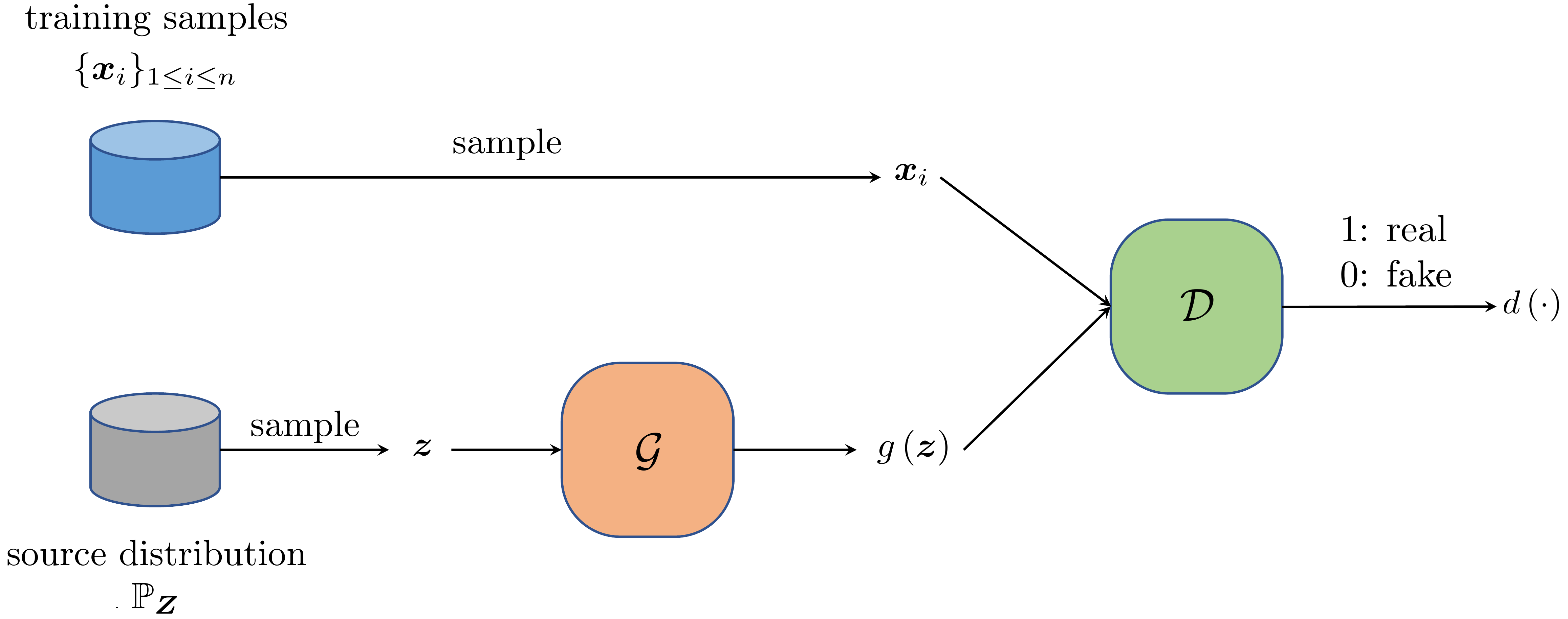}
\caption{GANs consist of two components, a generator $\mathcal{G}$ which generates fake samples and a discriminator $\mathcal{D}$ which differentiate the true ones from the fake ones. \label{fig:GAN}}
\end{figure}
\begin{equation}
\min_{\btheta_{\mathcal{G}}}\max_{\btheta_{\mathcal{D}}}
\qquad\mathbb{E}_{\bm{x}\sim\mathbb{P}_{\bm{X}}}\left[\log
\left(d\left(\bm{x}\right)\right)\right]
+\mathbb{E}_{\bm{z}\sim\mathbb{P}_{\bm{Z}}}\left[\log\left(1-d\left(\bg\left(\bm{z}\right)\right)\right)\right].\label{eq:GAN-original}
\end{equation}
Recall that $d(\bm{x})$ models the belief / probability that the
discriminator thinks that $\bm{x}$ is a true sample. Fix the parameters
$\btheta_{\mathcal{G}}$ and hence the generator $\mathcal{G}$ and
consider the inner maximization problem.
We can see that the goal
of the discriminator is to maximize its ability of differentiation.
Similarly, if we fix $\btheta_{\mathcal{D}}$ (and hence the discriminator),
the generator tries to generate more realistic images $\bg(\bm{z})$
to fool the discriminator.

\subsubsection{Density estimation view of GANs}
Let us now take a density-estimation view of GANs. Fixing the source distribution $\mathbb{P}_{\bm{Z}}$, any generator $\mathcal{G}$ induces a distribution $\mathbb{P}_{\mathcal{G}}$ over the space of images. Removing the restrictions on $d(\cdot)$, one can then rewrite (\ref{eq:GAN-original}) as
\begin{equation}\label{eq:GAN-new}
\min_{\mathbb{P}_{\mathcal{G}}}\max_{d(\cdot)}\qquad\mathbb{E}_{\bm{x}\sim\mathbb{P}_{\bm{X}}}\left[\log\left(d\left(\bm{x}\right)\right)\right]+\mathbb{E}_{\bm{x}\sim\mathbb{P}_{\mathcal{G}}}\left[\log\left(1-d\left(\bm{x}\right)\right)\right].
\end{equation}
Observe that the inner maximization problem
is solved by the likelihood ratio,~i.e.
\[
d^{*}\left(\bm{x}\right)=\frac{\mathbb{P}_{\bm{X}}\left(\bm{x}\right)}{\mathbb{P}_{\bm{X}}\left(\bm{x}\right)+\mathbb{P}_{\mathcal{G}}\left(\bm{x}\right)}.
\]
As a result, (\ref{eq:GAN-new}) can be simplified as
\begin{equation}
\min_{\mathbb{P}_{\mathcal{G}}}\qquad\text{JS}\left(\mathbb{P}_{\bm{X}}\;\|\;\mathbb{P}_{\mathcal{G}}\right)\label{eq:JS-min},
\end{equation}
where $\text{JS}(\cdot\|\cdot)$ denotes the Jensen--Shannon divergence
between two distributions
\[
\text{JS}\left(\mathbb{P}_{\bm{X}}\|\mathbb{P}_{\mathcal{G}}\right)=\frac{1}{2}\text{KL}\big(\mathbb{P}_{\bm{X}}\;\|\;\tfrac{\mathbb{P}_{\bm{X}}+\mathbb{P}_{\mathcal{G}}}{2}\big)+\frac{1}{2}\text{KL}\big(\mathbb{P}_{\mathcal{G}}\;\|\;\tfrac{\mathbb{P}_{\bm{X}}+\mathbb{P}_{\mathcal{G}}}{2}\big).
\]
In words, the vanilla GAN (\ref{eq:GAN-original}) seeks a density $\mathbb{P}_{\mathcal{G}}$ that is closest to $\mathbb{P}_{\bm{X}}$ in terms of the Jensen--Shannon divergence. This view allows to generalize GANs to other variants, by changing the distance metric. Examples include f-GAN \citep{nowozin2016f}, Wasserstein GAN (W-GAN) \citep{arjovsky2017wasserstein}, MMD GAN \citep{li2015generative}, etc. We single out the Wasserstein GAN (W-GAN) \citep{arjovsky2017wasserstein} to introduce due to its popularity. As the name suggests, it minimizes
the Wasserstein distance between $\mathbb{P}_{\bm{X}}$ and $\mathbb{P}_{\mathcal{G}}$:
\begin{equation}
\min_{\btheta_{\mathcal{G}}}\quad \text{WS}\left(\mathbb{P}_{\bm{X}}\|\mathbb{P}_{\mathcal{G}}\right)\;\;=\;\;\min_{\btheta_{\mathcal{G}}}\sup_{f:f\text{ 1-Lipschitz}}\mathbb{E}_{\bm{x}\sim\mathbb{P}_{\bm{X}}}
\left[f\left(\bm{x}\right)\right]-\mathbb{E}_{\bm{x}\sim
\mathbb{P}_{\mathcal{G}}}\left[f\left(\bm{x}\right)\right],\label{eq:WS-GAN}
\end{equation}
where $f(\cdot)$ is taken over all Lipschitz functions with coefficient 1.
Comparing W-GAN (\ref{eq:WS-GAN}) with the original formulation of GAN (\ref{eq:GAN-original}), one finds
that the Lipschitz function $f$ in (\ref{eq:WS-GAN}) corresponds
to the discriminator $\mathcal{D}$ in (\ref{eq:GAN-original}) in the sense that they
share similar objectives to differentiate the true distribution
$\mathbb{P}_{\bm{X}}$ from the fake one $\mathbb{P}_{\mathcal{G}}$. In the end, we would like to mention that GANs are more difficult to train than supervised deep learning models such as CNNs~\citep{salimans2016improved}. Apart from the training difficulty, how to evaluate GANs objectively and effectively is an ongoing research.

\section{Representation power: approximation theory}\label{sec:approx}
Having seen the building blocks of deep learning models in the previous sections, it is natural to ask: what is the benefits of composing multiple layers of nonlinear functions. In this section, we address this question from a approximation theoretical point of view. Mathematically, letting $\cH$ be the space of functions representable by neural nets (NNs),  how well can a function $f$ (with certain properties) be approximated by functions in $\cH$. We first revisit universal approximation theories, which are mostly developed for shallow neural nets (neural nets with a single hidden layer), and then provide recent results that demonstrate the benefits of depth in neural nets. Other notable works include Kolmogorov-Arnold superposition theorem~\citep{arnold2009functions, sprecher1965structure}, and circuit complexity for neural nets~\citep{parberry1994circuit}.


\subsection{Universal approximation theory for shallow NNs}
The universal approximation theories study the approximation of $f$ in a space~$\cF$ by a function represented by a one-hidden-layer neural net
\begin{equation}\label{def:HN}
g(\xx) = \sum_{j=1}^N c_j \sigma_*(\ww_j ^\top \xx - b_j),
\end{equation}
where $\sigma_*: \R \to \R$ is certain activation function and $N$ is the number of hidden units in the neural net. For different space $\cF$ and activation function $\sigma_*$, there are upper bounds and lower bounds on the approximation error $\| f - g \|$. See~\cite{pinkus1999approximation} for a comprehensive overview. Here we present representative results.

First, as $N \to \infty$, any continuous function $f$ can be approximated by some $g$ under mild conditions. Loosely speaking, this is because each component $\sigma_*(\ww_j ^\top \xx - b_j)$ behaves like a basis function 
and functions in a suitable space $\cF$ admits a basis expansion. Given the above heuristics, the next natural question is: what is the rate of approximation for a finite $N$?

Let us restrict the domain of $\xx$ to a unit ball $B^d$ in $\R^d$. For $p \in [1,\infty)$ and integer $m \ge 1$, consider the $L^p$ space and the Sobolev space with standard norms
\begin{align*}
 \| f \|_p = \Big[ \int_{B^n} | g(\xx) |^p \; d\xx  \Big]^{1/p}, \qquad \| f \|_{m,p} = \Big[ \sum_{0 \le |\kk| \le m} \| D^{\kk} f \|_p^p \Big]^{1/p},
\end{align*}
where $D^{\kk} f$ denotes partial derivatives indexed by $\kk \in \mathbb{Z}_+^d$. Let $\cF \triangleq \cF^m_p$ be the space of functions $f$ in the Sobolev space with $\| f \|_{m,p} \le 1$. Note that functions in~$\cF$ have bounded derivatives up to $m$-th order, and that smoothness of functions is controlled by $m$ (larger $m$ means smoother). Denote by $\cH_N$ the space of functions with the form \eqref{def:HN}. The following general upper bound is due to~\cite{mhaskar1996neural}.
\begin{thm}[Theorem~2.1 in \cite{mhaskar1996neural}]\label{thm:approx1}
Assume $\sigma_*: \R \to \R$ is such that $\sigma_*$ has arbitrary order derivatives in an open interval $I$, and that $\sigma_*$ is not a polynomial on $I$. Then, for any $p \in [1,\infty)$, $d \ge 2$, and integer $m \ge 1$,
\begin{equation*}
\sup_{f \in \cF^m_p} \inf_{g \in \cH_N^{\phantom{a}}} \| f - g \|_p \le C_{d,m,p}\, N^{-m/d},
\end{equation*}
where $C_{d,m,p}$ is independent of $N$, the number of hidden units.
\end{thm}
In the above theorem, the condition on $\sigma_*(\cdot)$ is mainly technical. This upper bound is useful when the dimension $d$ is not large. It clearly implies that the one-hidden-layer neural net is able to approximate any smooth function with enough hidden units. However, it is unclear how to find a good approximator $g$; nor do we have control over the magnitude of the parameters (huge weights are impractical). While increasing the number of hidden units $N$ leads to better approximation, the exponent $-m/d$ suggests the presence of the \emph{curse of dimensionality}. The following (nearly) matching lower bound is 
stated in~\cite{maiorov2000near}.
\begin{thm}[Theorem~5 in \cite{maiorov2000near}]\label{thm:approx2-2}
Let $p \ge 1$, $m \ge 1$ and $N \ge 2$. If the activation function is the standard sigmoid function $\sigma(t) = (1 + e^{-t})^{-1}$, then
\begin{equation}\label{ineq:approxlower2}
\sup_{f \in \cF^m_p} \inf_{g \in \cH_N^{\phantom{a}}} \| f - g \|_p \ge C'_{d,m,p}\, (N\log N)^{-m/d},
\end{equation}
where $C'_{d,m,p}$ is independent of $N$.
\end{thm}
Results for other activation functions are also obtained by~\cite{maiorov2000near}. Moreover, the term $\log N$ can be removed if we assume an additional continuity condition~\citep{mhaskar1996neural}.



For the natural space $\cF^m_p$ of smooth functions, the exponential dependence on $d$ in the upper and lower bounds may look unappealing. However,~\cite{barron1993universal} showed that for a different function space, there is a good dimension-free approximation by the neural nets. Suppose that a function $f: \mathbb{R}^{d} \mapsto \mathbb{R}$ has a Fourier representation
\begin{equation} \label{eq5.3}
f(\xx) = \int_{\R^{d}} e^{i \langle \bomega, \xx \rangle} \tilde f (\bomega)\; d\bomega,
\end{equation}
where $\tilde f (\bomega) \in \mathbb{C}$. Assume that $f(\bzero) = 0$ and that the following quantity is finite
\begin{equation}\label{def:Cf}
C_f = \int_{\R^{d}} \| \bomega \|_2 | \tilde f (\bomega) | \; d\bomega.
\end{equation}
\cite{barron1993universal} uncovers the following dimension-free approximation guarantee.
\begin{thm}[Proposition~1 in \cite{barron1993universal}]\label{thm:approx3}
Fix a $C>0$ and an arbitrary probability measure $\mu$ on the unit ball $B^d$ in $\R^d$. For every function $f$ with $C_f \le C$ and every $N \ge 1$, there exists some $g \in \cH_N$ such that
\begin{equation*}
\left[ \int_{B^d} ( f(\xx) - g(\xx))^2 \, \mu(d\xx) \right]^{1/2} \le \frac{2C}{\sqrt{N}}.
\end{equation*}
Moreover, the coefficients of $g$ may be restricted to satisfy $\sum_{j=1}^N |c_j| \le 2C$.
\end{thm}
The upper bound is now independent of the dimension $d$. 
However, $C_f$ may implicitly depend on $d$, as the formula in \eqref{def:Cf} involves an integration over $\R^{d}$ (so for some functions $C_f$ may depend exponentially on $d$). Nevertheless, this theorem does characterize an interesting function space with an improved upper bound. Details of the function space are discussed by~\cite{barron1993universal}. This theorem can be generalized; see~\cite{makovoz1996random} for an example.

To help understand why a dimensionality-free approximation holds, let us appeal to a heuristic argument given by Monte Carlo simulations. It is well-known that Monte Carlo approximation errors are independent of dimensionality in evaluation of high-dimensional integrals.  Let us generate $\{\bomega_j\}_{1\leq j \leq N}$ randomly from a given density $p(\cdot)$ in $\R^d$.  Consider the approximation to \eqref{eq5.3} by
\begin{equation} \label{eq5.4}
g_N(\xx) = \frac{1}{N} \sum_{j=1}^N c_j e^{i \langle \bomega_j, \xx \rangle}, \qquad c_j = \frac{\tilde f (\bomega_j)}{p(\bomega_j)}.
\end{equation}
Then, $g_N(\xx)$ is a one-hidden-layer neural network with $N$ units and the sinusoid activation function.  Note that $\E g_N(\xx) = f(\xx)$, where the expectation is taken with respect to randomness $\{\bomega_j\}$.  Now, by independence, we have
$$
    \E( g_N(\xx) - f(\xx))^2 = \frac{1}{N} \var(c_j e^{i \langle \bomega_j, \xx \rangle})\leq   \frac{1}{N} \E c_j^2,
$$
if $\E c_j^2 < \infty$.  Therefore, the rate is independent of the dimensionality $d$, though the constant can be.

\subsection{Approximation theory for multi-layer NNs}
The approximation theory for multilayer neural nets is less understood compared with neural nets with one hidden layer. Driven by the success of deep learning, there are many recent papers focusing on expressivity of deep neural nets. As studied by~\cite{telgarsky2016benefits, eldan2016power, mhaskar2016learning, poggio2017and, bauer2017deep, schmidt2017nonparametric, lin2017does,rolnick2017power}, deep neural nets excel at representing \textit{composition} of functions. This is perhaps not surprising, since deep neural nets are themselves defined by composing layers of  functions. Nevertheless, it points to a new territory rarely studied in statistics before. Below we present a result based on~\cite{lin2017does,rolnick2017power}.

Suppose that the inputs $\xx$ have a bounded domain $[-1,1]^d$ for simplicity. As before, let $\sigma_*: \R \to \R$ be a generic function, and $\bsigma_* = (\sigma_*, \cdots, \sigma_*)^\top$ be element-wise application of $\sigma_*$. Consider a neural net which is similar to \eqref{eq:fc} but with scaler output: $g(\xx) = \bW_\ell \bsigma_*(\cdots \bsigma_*(\bW_2 \bsigma_*(\bW_1 \xx))\cdots)$. A unit or neuron refers to an element of vectors $\bsigma_*(\bW_k \cdots \bsigma_*(\bW_2 \bsigma_*(\bW_1 \xx)) \cdots)$ for any $k=1,\ldots,\ell-1$. For a multivariate polynomial $p$, define $m_k(p)$ to be the smallest integer such that, for any $\epsilon > 0$, there exists a neural net $g(\xx)$ satisfying $\sup_\xx \left| p(\xx) - g(\xx) \right| < \epsilon$, with $k$ hidden layers (i.e., $\ell = k+1$) and no more than $m_k(p)$ neurons in total. Essentially, $m_k(p)$ is the minimum number of neurons required to approximate $p$ arbitrarily well.

\begin{thm}[Theorem~4.1 in \cite{rolnick2017power}]\label{thm:approx4}
Let $p(\xx)$ be a monomial $x_1^{r_1} x_2^{r_2} \cdots x_d^{r_d}$ with $q = \sum_{j=1}^d r_j$. Suppose that $\sigma_*$ has derivatives of order $2q$ at the origin, and that they are nonzero. Then,\\
(i) $m_1(p) = \prod_{j=1}^d (r_j + 1)$; \\
(ii) $\min_k m_k(p) \le \sum_{j=1}^d \left( 7 \lceil \log_2(r_j) \rceil + 4  \right)$.
\end{thm}

This theorem reveals a sharp distinction between shallow networks (one hidden layer) and deep networks. To represent a monomial function, a shallow network requires exponentially many neurons in terms of the dimension $d$, whereas linearly many neurons suffice for a deep network (with bounded $r_j$). The exponential dependence on $d$, as shown in Theorem~\ref{thm:approx4}(i), is resonant with the curse of dimensionality widely seen in many fields; see~\cite{donoho2000high}. One may ask: how does depth help? Depth circumvents this issue, at least for certain functions, by allowing us to represent function composition efficiently. Indeed, Theorem~\ref{thm:approx4}(ii) offers a nice result with clear intuitions: it is known that the product of two scalar inputs can be represented using $4$ neurons~\citep{lin2017does}, so by composing multiple products, we can express monomials with $O(d)$ neurons.

Recent advances in nonparametric regressions also support the idea that deep neural nets excel at representing composition of functions~\citep{bauer2017deep, schmidt2017nonparametric}. In particular,~\cite{bauer2017deep} considered the nonparametric regression setting where we want to estimate a function $\hat f_n(\xx)$ from i.i.d.~data $\mathcal{D}_n = \{ (y_i, \xx_i) \}_{1\leq i\leq n}$. If the true regression function $f(\xx)$ has certain hierarchical structure with intrinsic dimensionality\footnote{Roughly speaking, the true regression function can be represented by a tree where each node has at most $d^*$ children. See~\cite{bauer2017deep} for the precise definition.} $d^*$, then the error
\begin{equation*}
\E_{\mathcal{D}_n} \E_{\xx} \left| \hat f_n(\xx) - f(\xx) \right|^2
\end{equation*}
has an optimal minimax convergence rate $O(n^{-\frac{2q}{2q+d^*}})$, rather than the usual rate $O(n^{-\frac{2q}{2q+d}})$ that depends on the ambient dimension $d$. Here $q$ is the smoothness parameter. This provides another justification for deep neural nets: if data are truly hierarchical, then the quality of approximators by deep neural nets depends on the intrinsic dimensionality, which avoids the curse of dimensionality.
%

We point out that the approximation theory for deep learning is far from complete.
For example, in Theorem~\ref{thm:approx4}, the condition on $\sigma_*$ excludes the widely used ReLU activation function, there are no constraints on the magnitude of the weights (so they can be unreasonably large)
.


\section{Training deep neural nets }\label{sec:opt}
The \textit{existence} of a good function approximator in the NN function class does not explain why in practice we can easily \textit{find} them.
In this section, we introduce standard methods, namely \emph{stochastic gradient descent} (SGD) and its variants, to train deep neural networks (or to find such a good approximator). As with many statistical machine learning tasks, training DNNs follows the \emph{empirical risk minimization} (ERM) paradigm which solves the following optimization problem
\begin{equation}
\text{minimize}_{\btheta\in\mathbb{R}^{p}}\qquad\ell_{n}\left(\btheta\right)\triangleq\frac{1}{n}\sum_{i=1}^{n}\mathcal{L}\left(f\left(\bm{x}_{i};\btheta\right),y_{i}\right).\label{eq:ERM_for_DL}
\end{equation}
Here $\mathcal{L}(f(\bm{x}_{i};\btheta),y_{i})$ measures the discrepancy between the prediction $f(\bm{x}_{i};\btheta)$ of the neural network and the true label $y_{i}$. Correspondingly, denote by $\ell(\btheta) \triangleq \mathbb{E}_{(\bm{x},y)\sim\mathcal{D}}[\mathcal{L}(f(\bm{x};\btheta),\bm{y})]$ the out-of-sample error, where $\mathcal{D}$ is the joint distribution over $(y, \bm{x})$.  Solving ERM~(\ref{eq:ERM_for_DL}) for deep neural nets faces various challenges that roughly fall into the following three categories.
\begin{itemize}
\item \emph{Scalability and nonconvexity.} Both the sample size $n$ and the number of parameters $p$ can be huge for modern deep learning applications, as we have seen in Table~\ref{tab:intro}.
Many optimization algorithms are not practical due to the computational costs and memory constraints. What is worse, the empirical loss function $\ell_{n}(\bm{\theta})$ in deep learning is often nonconvex. It is \emph{a priori }not clear whether an optimization algorithm can drive the empirical loss (\ref{eq:ERM_for_DL}) small.

\item \emph{Numerical stability.} With a large number of layers in DNNs, the magnitudes of the hidden nodes can be drastically different, which may result in the ``exploding gradients'' or ``vanishing gradients'' issue during the training process. This is because the recursive relations across layers often lead to exponentially increasing$\,$/$\,$decreasing values in both forward passes and backward passes.

\item \emph{Generalization performance.} Our ultimate goal is to find a parameter $\hat {\bm{\theta}}$ such that the out-of-sample error $\ell(\hat \btheta)$ is small. 
However, in the over-parametrized regime where $p$ is much larger than $n$, the underlying neural network has the potential to fit the training data perfectly while performing poorly on the test data. To avoid this overfitting issue, proper regularization, whether explicit or implicit, is needed in the training process for the neural nets to generalize.
\end{itemize}

In the following three subsections, we discuss practical solutions$\,$/$\,$proposals to address these challenges.

\subsection{Stochastic gradient descent \label{sec:stochastic-opt}}

Stochastic gradient descent (SGD)~\citep{robbins1951stochastic} is by far the most popular optimization algorithm to solve ERM~(\ref{eq:ERM_for_DL})
for large-scale problems. It has the following simple update rule:
\begin{equation}
\btheta^{t+1}=\btheta^{t}-\eta_{t}G(\bm{\theta}^{t})\qquad\text{with}\qquad G\left(\bm{\theta}^{t}\right)=\nabla\mathcal{L}\left(f\left(\bm{x}_{i_{t}};\btheta^{t}\right),y_{i_{t}}\right)\label{eq:SGD_DL}
\end{equation}
for $t=0,1,2,\ldots$, where $\eta_{t}>0$ is the step size (or learning rate), $\btheta^{0}\in\mathbb{R}^{p}$ is an initial point and $i_{t}$ is chosen randomly from $\{1,2,\cdots, n\}$. It is easy to verify that $G(\bm{\theta}^{t})$ is an unbiased estimate of $\nabla\ell_{n}(\bm{\theta}^{t})$.
The advantage of SGD is clear: compared with gradient descent, which goes over the entire dataset in every update, SGD uses a single example in each update and hence is considerably more efficient in terms of both computation and memory (especially in the first few iterations).

Apart from practical benefits of SGD, how well does SGD perform theoretically in terms of minimizing $\ell_{n}(\btheta)$? We begin with the convex case, i.e., the case where the loss function is convex w.r.t.~$\bm{\theta}$. It is well understood in literature that with proper choices of the step sizes $\{\eta_{t}\}$, SGD is guaranteed to achieve both \emph{consistency} and \emph{asymptotic normality}.
\begin{itemize}
\item {\emph{Consistency}.} If $\ell(\btheta)$ is a strongly convex function\footnote{For results on consistency and asymptotic normality, we consider the case where in each step of SGD, the stochastic gradient is computed using a fresh sample $(y, \bm{x})$ from $\mathcal{D}$. This allows to view SGD as an optimization algorithm to minimize the population loss $\ell(\btheta)$.}, then under some mild conditions\footnote{One example of such condition can be constraining the second moment of the gradients: $\E\left[\|\nabla\mathcal{L}\left(\bm{x}_{i},y_{i};\btheta^{t}\right)\|_{2}^{2}\right]\le C_{1}+C_{2}\|\btheta^{t}-\btheta^{*}\|_{2}^{2}$ for some $C_{1},C_{2}>0$. See~\cite{bottou1998online} for details.}, learning rates that satisfy
\begin{equation}\label{cond:lr}
\sum_{t=0}^{\infty}\eta_{t}=+\infty\qquad\text{and}\qquad\sum_{t=0}^{\infty}\eta_{t}^{2}<+\infty
\end{equation}
guarantee almost sure convergence to the unique minimizer $\btheta^* \triangleq \argmin_{\btheta} \ell(\btheta)$, i.e., $\btheta^{t}\xrightarrow{\mbox{a.s.}}\btheta^{*}$ as $t\to\infty$~\citep{robbins1951stochastic,kiefer1952stochastic,bottou1998online,kushner2003stochastic}. The requirements in \eqref{cond:lr} can be viewed from the perspective of bias-variance tradeoff: the first condition ensures that the iterates can reach the minimizer (controlled bias), and the second ensures that stochasticity does not prevent convergence (controlled variance).

\item {\emph{Asymptotic normality}.} It is proved by~\cite{polyak1979adaptive} that for robust linear regression with fixed dimension $p$, under the choice $\eta_{t}=t^{-1}$, $\sqrt{t}\,(\btheta^{t}-\btheta^{*})$ is asymptotically normal under some regularity conditions (but $\btheta^{t}$ is not asymptotically efficient in general). Moreover, by averaging the iterates of SGD,~\cite{polyak1992acceleration} proved that even with a \emph{larger} step size $\eta_{t}\propto t^{-\alpha},\alpha\in(1/2,1)$, the averaged iterate $\bar{\btheta}^{t}=t^{-1}\sum_{s=1}^{t}\btheta^{s}$ is asymptotic efficient for robust linear regression. These strong results show that SGD with averaging performs as well as the MLE asymptotically, in addition to its computational efficiency. 
\end{itemize}

These classical results, however, fail to explain the effectiveness of SGD when dealing with nonconvex loss functions in deep learning. Admittedly, finding global minima of nonconvex functions is computationally infeasible in the worst case. Nevertheless, recent work~\citep{allen2018convergence,du2018gradient} bypasses the worst case scenario by focusing on losses incurred by over-parametrized deep learning models. In particular, they show that (stochastic) gradient descent converges linearly towards the \emph{global }minimizer of $\ell_{n}(\btheta)$ as long as the neural network is sufficiently \emph{over-parametrized}. This phenomenon is formalized below.

\begin{thm}[Theorem~2 in \citealp{allen2018convergence}]Let $\{(y_i, \bm{x}_{i})\}_{1 \leq i \leq n}$
be a training set satisfying $\min_{i,j:i\neq j}\|\bm{x}_{i}-\bm{x}_{j}\|_{2}\geq\delta>0$. Consider fitting the data using a feed-forward neural network~(\ref{model:1}) with
ReLU activations. Denote by $L$ (resp.~$W$) the depth (resp.~width) of the network. Suppose that the neural network is sufficiently over-parametrized, i.e.,
\begin{equation}
W\gg\mathsf{poly}\left(n,L,\frac{1}{\delta}\right), \label{eq:overparametrize}
\end{equation}
where $\mathsf{poly}$ means a polynomial function. Then with high probability, running SGD~(\ref{eq:SGD_DL}) with \emph{certain
random initialization} and properly chosen step sizes yields $\ell_{n}(\bm{\theta}^{t})\leq\varepsilon$
in $t\asymp\log\frac{1}{\varepsilon}$ iterations. \end{thm}

Two
notable features are worth mentioning:~(1) first, the network under
consideration is sufficiently over-parametrized (cf.~(\ref{eq:overparametrize})) in which the number of parameters is \emph{much} larger than the number of samples,
and (2) one needs to initialize the weight matrices to be in near-isometry such that the magnitudes of the hidden nodes do not blow up or vanish. In a nutshell, \emph{over-parametrization}
and \emph{random initialization} together ensure that the loss function
(\ref{eq:ERM_for_DL}) has a benign landscape\footnote{In~\cite{allen2018convergence},
the loss function $\ell_{n}(\btheta)$ satisfies the PL condition.} around the initial
point, which in turn implies fast convergence of SGD iterates.

There are certainly other challenges for vanilla SGD to train deep neural nets: (1) training algorithms are often implemented in GPUs, and therefore it
is important to tailor the algorithm to the infrastructure, (2) the
vanilla SGD might converge very slowly for deep neural networks,
albeit good theoretical guarantees for well-behaved problems, and (3) the learning
rates $\{\eta_{t}\}$ can be difficult to tune in practice. To address
the aforementioned challenges, three important variants of SGD, namely
\emph{mini-batch SGD, momentum-based SGD}, and \emph{SGD with adaptive
learning rates }are introduced.

\subsubsection{Mini-batch SGD}

Modern computational infrastructures (e.g., GPUs)
can evaluate the gradient on a number (say 64) of examples as efficiently
as evaluating that on a single example. To utilize this advantage, mini-batch SGD
with batch size $K\geq1$ forms the stochastic gradient through $K$
random samples:
\begin{equation}
\bm{\theta}^{t+1}=\bm{\theta}^{t}-\eta_{t}G(\bm{\theta}^{t})\qquad\text{with}\qquad G(\bm{\theta}^{t})=\frac{1}{K}\sum_{k=1}^{K}\nabla\mathcal{L}\big(f\big(\bm{x}_{i_{t}^{k}};\btheta^{t}\big),y_{i_{t}^{k}}\big),\label{eq:mini-SGD}
\end{equation}
where for each $1\leq k\leq K$, $i_{t}^{k}$ is sampled uniformly
from $\{1,2,\cdots,n\}$. Mini-batch SGD, which is an ``interpolation''
between gradient descent and stochastic gradient descent, achieves
the best of both worlds: (1) using $1\ll K\ll n$ samples to estimate
the gradient, one effectively reduces the variance and hence accelerates
the convergence, and (2) by taking the batch size $K$ appropriately (say
64 or 128), the stochastic gradient $G(\bm{\theta}^{t})$ can be efficiently
computed using the matrix computation toolboxes on GPUs.

\subsubsection{Momentum-based SGD}
While mini-batch SGD forms the foundation
of training neural networks, it can sometimes be slow to converge
due to its oscillation behavior~\citep{sutskever2013importance}. Optimization community has long investigated
how to accelerate the convergence of gradient descent,
which results in a beautiful technique called \emph{momentum methods}
\citep{polyak1964some,nesterov1983method}. Similar to gradient descent with moment, \emph{momentum-based SGD}, instead of moving the iterate
$\bm{\theta}^{t}$ in the direction of the current stochastic gradient $G(\btheta^t)$, smooth the past (stochastic) gradients $\{G(\btheta^t)\}$ to stabilize the update directions. Mathematically,
let $\bm{v}^{t}\in\mathbb{R}^{p}$ be the direction of update in the
$t$th iteration, i.e.,
\[
\bm{\theta}^{t+1}=\bm{\theta}^{t}-\eta_{t}\bm{v}^{t}.
\]
Here $\bm{v}^{0}=G(\bm{\theta}^{0})$ and for $t=1,2,\cdots$
\begin{equation} \label{eq6.6}
\bm{v}^{t}=\rho\bm{v}^{t-1}+G(\bm{\theta}^{t})
\end{equation}
with $0<\rho<1$. A typical choice of $\rho$ is 0.9. Notice
that $\rho=0$ recovers the mini-batch SGD (\ref{eq:mini-SGD}),
where no past information of gradients is used. A simple unrolling
of $\bm{v}^{t}$ reveals that $\bm{v}^{t}$ is actually an exponential
averaging of the past gradients, i.e., $\bm{v}^{t}=\sum_{j=0}^{t}\rho^{t-j}G(\bm{\theta}^{j}).$
Compared with vanilla mini-batch SGD, the inclusion of the momentum
``smoothes'' the oscillation direction and accumulates the persistent
descent direction. We want to emphasize that theoretical justifications of momentum in the \emph{stochastic} setting is not fully understood~\citep{kidambi2018insufficiency, jain2017accelerating}.


\subsubsection{SGD with adaptive learning rates}

In optimization, \emph{preconditioning} is often used to accelerate first-order optimization algorithms. In principle, one can apply this to SGD, which yields the following update rule:
\begin{equation}
\bm{\theta}^{t+1}=\bm{\theta}^{t}-\eta_{t}\bm{P}_{t}^{-1}G(\bm{\theta}^{t})\label{eq:precondition-SGD}
\end{equation}
with $\bm{P}_t \in \mathbb{R}^{p\times p}$ being a preconditioner at the $t$-th step. Newton's method can be viewed as one type of preconditioning where $\bm{P}_{t} = \nabla^2 \ell(\btheta^t)$. The advantages of preconditioning are two-fold: first, a good preconditioner reduces the condition number by changing the local geometry to be more homogeneous, which is amenable to fast convergence; second, a good preconditioner frees practitioners from laboring tuning of the step sizes, as is the case with Newton's method. AdaGrad, an adaptive gradient method proposed by~\cite{duchi2011adaptive}, builds a preconditioner $\bm{P}_{t}$ based on information of the past gradients:
\begin{equation}
\bm{P}_{t}= \Big \{ \mathsf{diag}\Big(\sum_{j=0}^{t}G\left(\bm{\theta}^{t}\right)G
\left(\bm{\theta}^{t}\right)^{\top}\Big) \Big \}^{1/2} \label{eq:adagrad-precondition}.
\end{equation}
Since we only require the diagonal part, this preconditioner (and its inverse) can be efficiently computed in practice. In addition, investigating~(\ref{eq:precondition-SGD}) and~(\ref{eq:adagrad-precondition}), one can see that AdaGrad adapts to the importance
of each coordinate of the parameters by setting smaller learning rates
for frequent features, whereas larger learning rates for those infrequent
ones. In practice, one adds a small quantity $\delta > 0$ (say $10^{-8}$) to the diagonal entries to avoid singularity (numerical underflow). A notable drawback of AdaGrad is that the effective learning rate vanishes quickly along the learning process. This
is because the historical sum of the gradients can only
increase with time. RMSProp~\citep{hinton2012neural} is a popular
remedy for this problem which incorporates the idea of exponential averaging:
\begin{equation}
\bm{P}_{t}= \Big \{ \mathsf{diag}\Big(\rho \bm{P}_{t-1} +
(1-\rho)G\left(\bm{\theta}^{t}\right)G\left(\bm{\theta}^{t}\right)^{\top}
\Big) \Big \}^{1/2} \label{eq:rmsprop-precondition}.
\end{equation}
Again, the decaying parameter $\rho$ is usually set to be $0.9$.
Later, Adam~\citep{kingma2014adam, reddi2018convergence} combines
the momentum method and adaptive learning rate and becomes the default training algorithms in many deep learning applications.

\subsection{Easing numerical instability}\label{sec:batch-norm}


For very deep neural networks or RNNs with long dependencies, training difficulties often arise when the values of nodes have different magnitudes or when the gradients ``vanish'' or ``explode'' during back-propagation. Here we discuss three partial solutions to alleviate this problem.

\subsubsection{ReLU activation function}

One useful characteristic of the ReLU function is that its derivative is either $0$ or $1$, and the derivative remains $1$ even for a large input. This is in sharp contrast with the standard sigmoid function $(1 + e^{-t})^{-1}$ which results in a very small derivative when inputs have large magnitude. The consequence of small derivatives across many layers is that gradients tend to be ``killed'', which means that gradients become approximately zero in deep nets. 

The popularity of the ReLU activation function and its variants (e.g., leaky ReLU) is largely attributable to the above reason. It has been well observed that the ReLU activation function has superior training performance over the sigmoid function~\citep{krizhevsky2012imagenet, maas2013rectifier}.

\subsubsection{Skip connections}

We have introduced skip connections in Section~\ref{sec:skip}. Why are skip connections helpful for reducing numerical instability? This structure does not introduce a larger function space, since the identity map can be also represented with ReLU activations: $\xx = \bsigma(\xx) - \bsigma(-\xx)$.

One explanation is that skip connections bring ease to the training$\,$/$\,$optimization process. Suppose that we have a general nonlinear function $\bF(\xx_\ell; \btheta_{\ell})$. With a skip connection, we represent the map as $\xx_{\ell+1} = \xx_\ell + \bF(\xx_\ell; \btheta_{\ell})$ instead. Now the gradient $\partial \xx_{\ell+1} / \partial \xx_{\ell}$ becomes
\begin{equation}\label{eq:skipgrad}
\frac{\partial \xx_{\ell+1}}{\partial \xx_{\ell}} = \bI + \frac{\partial \bF(\xx_\ell; \btheta_{\ell})}{\partial \xx_\ell} \qquad \text{instead of} \qquad \frac{\partial \bF(\xx_\ell; \btheta_{\ell})}{\partial \xx_\ell},
\end{equation}
where $\bI$ is an identity matrix. By the chain rule, gradient update requires computing products of many components, e.g., $\frac{\partial \xx_L}{\partial \xx_1} = \prod_{\ell=1}^{L-1} \frac{\partial \xx_{\ell+1}}{\partial \xx_\ell}$, so it is desirable to keep the spectra (singular values) of each component $\frac{\partial \xx_{\ell+1}}{\partial \xx_\ell}$ close to $1$. In neural nets, with skip connections, this is easily achieved if the parameters have small values; otherwise, this may not be achievable even with careful initialization and tuning. Notably, training neural nets with hundreds of layers is possible with the help of skip connections.


\subsubsection{Batch normalization}

Recall that in regression analysis, one often standardizes the design matrix so that the features have zero mean and unit variance. Batch normalization extends this standardization procedure from the input layer to all the hidden layers. Mathematically, fix a mini-batch of input data $\{(\bm{x}_{i},y_{i})\}_{i\in \mathcal{B}}$, where $\mathcal{B} \subset [n]$. Let $\bm{h}_{i}^{(\ell)}$ be the feature of the $i$-th example in the $\ell$-th layer ($\ell=0$ corresponds to the input $\bm{x}_{i}$). The batch normalization layer computes the normalized version of $\bm{h}_{i}^{(\ell)}$ via the following steps:
\begin{align*}
\bm{\mu} & \triangleq \frac{1}{\left|\mathcal{B}\right|}\sum_{i\in \mathcal{B}}\bm{h}_{i}^{(\ell)},\qquad \bm{\sigma}^{2}  \triangleq\frac{1}{\left|\mathcal{B}\right|}\sum_{i\in \mathcal{B}}\big(\bm{h}_{i}^{(\ell)}-\bm{\mu}\big)^{2} \qquad\text{and}\qquad \bm{h}_{i,\text{norm}}^{(l)}  \triangleq\frac{\bm{h}_{i}^{(\ell)}-\bm{\mu}}{\bm{\sigma}}.
\end{align*}
Here all the operations are element-wise. In words, batch normalization computes the z-score for each feature over the mini-batch $\mathcal{B}$ and use that as inputs to subsequent layers. To make it more versatile, a typical batch normalization layer has two additional learnable parameters $\bm{\gamma}^{(\ell)}$ and $\bm{\beta}^{(\ell)}$ such that
\[
\bm{h}_{i,\text{new}}^{(l)}=\bm{\gamma}^{(l)}\odot\bm{h}_{i,\text{norm}}^{(l)}+\bm{\beta}^{(l)}.
\]
Again $\odot$ denotes the element-wise multiplication. As can be seen, $\bm{\gamma}^{(\ell)}$ and $\bm{\beta}^{(\ell)}$ set the new feature $\bm{h}_{i \text{new}}^{(l)}$ to have mean $\bm{\beta}^{(\ell)}$ and standard deviation $\bm{\gamma}^{(\ell)}$. The introduction of batch normalization makes the training of neural networks much easier and smoother. More importantly, it allows the neural nets to perform well over a large family of hyper-parameters including the number of layers, the number of hidden units, etc. At test time, the batch normalization layer needs more care. For brevity we omit the details and refer to~\cite{ioffe2015batch}.




\subsection{Regularization techniques} \label{sec:regularization}
So far we have focused on training techniques to drive the empirical loss (\ref{eq:ERM_for_DL}) small efficiently. Here we proceed to discuss common practice to improve the generalization power of trained neural nets.

\subsubsection{Weight decay} \label{sec:weight}
One natural regularization idea is to add an $\ell_{2}$
penalty to the loss function. This regularization technique is known
as the weight decay in deep learning. We have seen one example in
\eqref{eq:regloss}. For general deep neural nets, the loss to optimize
is $\ell_n^{\lambda}(\btheta)=\ell_{n}(\btheta)+r_{\lambda}(\btheta)$ where 
\[
r_{\lambda}(\btheta)=\lambda\sum_{\ell=1}^{L}\sum_{j,j'}\big[W_{j,j'}^{(\ell)}\big]^{2}.
\]
Note that the bias (intercept) terms are not penalized. If $\ell_{n}(\btheta)$
is a least square loss, then regularization with weight decay gives
precisely ridge regression. The penalty $r_{\lambda}(\btheta)$ is a smooth
function and thus it can be also implemented efficiently with back-propagation.

\subsubsection{Dropout} \label{sec:dropout}

Dropout, introduced by~\cite{hinton2012improving}, prevents overfitting by randomly dropping out subsets of features
during training. Take the $l$-th layer of the feed-forward neural
network as an example. Instead of propagating all the features in $\bm{h}^{(\ell)}$ for later
computations, dropout randomly omits some of its entries by
\[
\bm{h}_{\text{drop}}^{(\ell)}=\bm{h}^{(\ell)}\odot\mathsf{mask}^{\ell},
\]
where $\odot$ denotes element-wise multiplication as before, and $\mathsf{mask}^{\ell}$ is a vector of Bernoulli variables with success probability $p$. It is sometimes useful to rescale the features $\bm{h}_{\text{inv drop}}^{(\ell)}=\bm{h}_{\text{drop}}^{(\ell)}/p$, which is called \textit{inverted dropout}. During training, $\mathsf{mask}^{\ell}$ are i.i.d. vectors across mini-batches and layers. However, when testing on fresh samples, dropout is disabled
and the original features $\hh^{(\ell)}$ are used to compute the
output label $y$. It has been nicely shown by~\cite{wager2013dropout}
that for generalized linear models, dropout serves as adaptive
regularization. In the simplest case of linear regression, it is equivalent
to $\ell_{2}$ regularization. Another possible way to understand
the regularization effect of dropout is through the lens of bagging~\citep{deeplearningbook}.
Since different mini-batches has different masks, dropout can be viewed
as training a large ensemble of classifiers at the same time, with
a further constraint that the parameters are shared. 
Theoretical justification remains elusive.
%


\subsubsection{Data augmentation}\label{sec:aug}
Data augmentation is a technique of enlarging the dataset when we have knowledge about invariance structure of data. It implicitly increases the sample size and usually regularizes the model effectively. For example, in image classification, we have strong prior knowledge about what invariance properties a good classifier should possess. The label of an image should not be affected by translation, rotation, flipping, and even crops of the image. Hence one can augment the dataset by randomly translating, rotating and cropping the images in the original dataset.

Formally, during training we want to minimize the loss $\ell_{n}(\btheta)=\sum_{i}\cL(f(\xx_{i};\btheta),y_{i})$
w.r.t.~parameters $\btheta$, and we know a priori that certain transformation
$T\in\mathcal{T}$ where $T:\R^{d}\to\R^{d}$ (e.g., affine transformation)
should not change the category$\,$/$\,$label of a training sample. In principle,
if computation costs were not a consideration, we could convert this
knowledge to a constraint $f_{\btheta}(T\xx_{i})=f_{\btheta}(\xx_{i}),\forall\,T\in\mathcal{T}$
in the minimization formulation. Instead of solving a constrained
optimization problem, data augmentation enlarges the training dataset
by sampling $T\in\mathcal{T}$ and generating new data $\{(T\xx_{i},y_{i})\}$.
In this sense, data augmentation induces invariance properties through sampling, which results in a much bigger dataset than the original one.




%

\section{Generalization power}\label{sec:generalization}

Section~\ref{sec:opt} has focused on the in-sample$\,$/$\,$ training
error obtained via SGD, but this alone does not guarantee good performance with respect to the
out-of-sample$\,$/$\,$test error. The gap between the in-sample
error and the out-of-sample error, namely the \emph{generalization
gap}, has been the focus of statistical learning theory since its
birth; see \cite{shalev2014understanding} for an excellent introduction
to this topic.


While understanding the generalization power of deep neural nets is difficult~\cite{zhang2016understanding,recht2018cifar}, we sample recent endeavors in this section. From a high level point of
view, these approaches can be divided into two categories, namely \emph{algorithm-independent controls }and \emph{algorithm-dependent
control}s. More specifically, algorithm-independent controls focus
solely on bounding the \emph{complexity }of the function class represented
by certain deep neural networks. In contrast, algorithm-dependent
controls take into account the algorithm (e.g., SGD) used
to train the neural network.

\subsection{Algorithm-independent controls: uniform convergence}

The key to algorithm-independent controls is the notion of \emph{complexity
}of the function class parametrized by certain neural networks. Informally,
as long as the complexity is not too large, the generalization gap
of \emph{any} function in the function class is well-controlled. However,
the standard complexity measure (e.g., VC dimension \citep{vapnik1971uniform})
is at least proportional to the number of weights
in a neural network \citep{anthony2009neural, shalev2014understanding}, which fails to
explain the practical success of deep learning. The caveat here is
that the function class under consideration is \emph{all} the functions realized
by certain neural networks, with \emph{no} restrictions on the size
of the weights at all. On the other hand, for the class of linear
functions with bounded norm, i.e., $\{\bm{x}\mapsto\bm{w}^{\top}\bm{x}\,|\,\|\bm{w}\|_{2}\leq M\}$,
it is well understood that the complexity of
this function class (measured in terms of the empirical Rademacher complexity) with respect to a random sample $\{\bm{x}_{i}\}_{1 \leq i \leq n}$
is upper bounded by $\max_{i}\|\bm{x}_{i}\|_{2}M/\sqrt{n}$, which
is independent of the number of parameters in $\bm{w}$. This motivates
researchers to investigate the complexity of \emph{norm-controlled}
deep neural networks\footnote{Such attempts have been made in the seminal work~\cite{bartlett1998sample}.} \citep{neyshabur2015norm,NIPS2017_7204,golowich2017size,li2018tighter}.
Setting the stage, we introduce a few necessary notations and facts.
The key object under study is the function class parametrized by the
following fully-connected neural network with depth~$L$:
\begin{equation}
\mathcal{F}_{L}\triangleq\left\{ \bm{x}\mapsto\bm{W}_{L}\bsigma\left(\bm{W}_{L-1}\bsigma\left(\cdots\bm{W}_{2}\bsigma\left(\bm{W}_{1}\bm{x}\right)\right)\right)\,\big|\,\left(\bm{W}_{1},\cdots,\bm{W}_{L}\right)\in\mathcal{W}\right\}.\label{eq:function-class-ffn}
\end{equation}
Here $(\bm{W}_{1},\bm{W}_{2},\cdots,\bm{W}_{L})\in\mathcal{W}$
represents a certain constraint on the parameters. For instance, one
can restrict the Frobenius norm of each parameter $\bm{W}_{l}$ through
the constraint $\|\bm{W}_{l}\|_{\mathrm{F}}\leq M_{\mathrm{F}}(l)$,
where $M_{\mathrm{F}}(l)$ is some positive quantity. With regard
to the complexity measure, it is standard to use \emph{Rademacher
complexity} to control the capacity of the function class of interest.

\begin{definition}[Empirical Rademacher complexity] The empirical
Rademacher complexity of a function class~$\mathcal{F}$ w.r.t.~a dataset $S\triangleq\{\bm{x}_{i}\}_{1 \leq i \leq n}$  is
defined as
\begin{equation}
\mathcal{R}_{S}\left(\mathcal{F}\right)=\mathbb{E}_{\bm{\varepsilon}}\Big[\sup_{f\in\mathcal{F}}\frac{1}{n}\sum_{i=1}^{n}\varepsilon_{i}f\left(\bm{x}_{i}\right)\Big],\label{eq:empirical-rademacher-complexity}
\end{equation}
where $\bm{\varepsilon}\triangleq(\varepsilon_{1},\varepsilon_{2},\cdots,\varepsilon_{n})$
is composed of i.i.d.~Rademacher random variables, i.e., $\mathbb{P}(\varepsilon_{i}=1)=\mathbb{P}(\varepsilon_{i}=-1)=1/2$.
\end{definition}

In words, Rademacher complexity measures
the ability of the function class to fit the random noise represented
by $\bm{\varepsilon}$. Intuitively, a function class with a larger Rademacher
complexity is more prone to overfitting. We now formalize the connection between the empirical Rademacher complexity
and the out-of-sample error; see Chapter 24 in~\cite{shalev2014understanding}.

\begin{thm}Assume that for all $f\in\mathcal{F}$
and all $(y,\bm{x})$ we have $|\mathcal{L}(f(\bm{x}),y)|\leq1$.
In addition, assume that for any fixed $y$, the univariate function $\mathcal{L}(\cdot,y)$ 
is Lipschitz with constant 1. Then with probability at least $1-\delta$ over the sample $S\triangleq\{(y_{i},\bm{x}_{i})\}_{1\leq i\leq n}\overset{\mathrm{i.i.d.}}{\sim}\mathcal{D}$, one has for all $f\in\mathcal{F}$
\[
\underbrace{\vphantom{\frac{1}{n}\sum_{i=1}^{n}}\mathbb{E}_{(y,\bm{x})\sim \mathcal{D}}\left[\mathcal{L}\left(f(\bm{x}), y\right)\right]}_{\mathrm{out}\text{-}\mathrm{of}\text{-}\mathrm{sample\;error}}\leq\underbrace{\frac{1}{n}\sum_{i=1}^{n}\mathcal{L}\left(f(\bm{x}_{i}),y_{i}\right)}_{\mathrm{in}\text{-}\mathrm{sample\;error}}+2\mathcal{R}_{S}\left(\mathcal{F}\right)+4\sqrt{\frac{\log\left(4/\delta\right)}{n}}.
\]
\end{thm}
In English, the generalization gap of any function $f$ that lies in $\mathcal{F}$ is well-controlled as long as the Rademacher complexity of $\mathcal{F}$ is not too large. With this connection in place, we single out the
following complexity bound.

\begin{thm}[Theorem 1 in~\citep{golowich2017size}]Consider the
function class $\mathcal{F}_{L}$ in~(\ref{eq:function-class-ffn}),
where each parameter $\bm{W}_{l}$ has Frobenius norm at most $M_{\mathrm{F}}(l)$.
Further suppose that the element-wise activation function $\sigma(\cdot)$
is $1$-Lipschitz and positive-homogeneous (i.e., $\sigma(c\cdot x)=c\sigma(x)$
for all $c\geq0$). Then the empirical Rademacher complexity~(\ref{eq:empirical-rademacher-complexity})
w.r.t. $S\triangleq\{\bm{x}_{i}\}_{1 \leq i \leq n}$ satisfies
\begin{equation}
\mathcal{R}_{S}\left(\mathcal{F}_{L}\right)\leq\max_{i}\|\bm{x}_{i}\|_{2}\cdot\frac{4\sqrt{L}\prod_{l=1}^{L}M_{\mathrm{F}}(l)}{\sqrt{n}}.\label{eq:rademacher-nn}
\end{equation}
\end{thm}The upper bound of the empirical Rademacher complexity
(\ref{eq:rademacher-nn}) is in a similar vein to that of linear functions
with bounded norm, i.e., $\max_{i}\|\bm{x}_{i}\|_{2}M/\sqrt{n}$,
where $\sqrt{L}\prod_{l=1}^{L}M_{\mathrm{F}}(l)$ plays the role of
$M$ in the latter case. Moreover, ignoring the term $\sqrt{L}$,
the upper bound (\ref{eq:rademacher-nn}) does not depend on the size
of the network in an explicit way  if $M_F(l)$ sharply concentrates around $1$. This reveals that the capacity
of the neural network is well-controlled, regardless of the number
of parameters, as long as the Frobenius norm of the parameters is
bounded. Extensions to other norm constraints, e.g., spectral norm
constraints, path norm constraints have been considered by \cite{neyshabur2015norm,NIPS2017_7204,li2018tighter,klusowski2016risk, E19}.
This line of work improves upon traditional capacity analysis of neural
networks in the over-parametrized setting, because the upper bounds derived are often size-independent.
Having said this, two important remarks are in order: (1) the upper bounds (e.g., $\prod_{l=1}^{L}M_{\mathrm{F}}(l)$)
involve implicit dependence on the size of the weight matrix and
the depth of the neural network, which is hard to characterize; (2) the upper bound on the Rademacher complexity offers a uniform bound
over all functions in the function class, which is a pure statistical
result. However, it stays silent about how and why standard training
algorithms like SGD can obtain a function whose parameters have small
norms.

\subsection{Algorithm-dependent controls}

In this subsection, we bring computational thinking into statistics and investigate the role of algorithms in the generalization power of deep learning. The consideration of algorithms is quite natural and well motivated: (1) local/global minima reached by different algorithms can exhibit totally different generalization behaviors due to extreme nonconvexity, which marks a huge difference from traditional models, (2) the \emph{effective }capacity of neural nets is possibly not large, since a particular algorithm does not explore the entire parameter space.

These demonstrate the fact that on top of the complexity of the function class,
the inherent property of the algorithm we use plays an important role in the generalization ability of deep learning. In what follows, we survey three different ways to obtain upper bounds on the generalization errors by exploiting properties of the algorithms.
\subsubsection{Mean field view of neural nets} As we have emphasized, modern deep learning models are highly over-parametrized. 
A line of work~\citep{mei2018mean,sirignano2018mean,rotskoff2018neural,chizat2018global,mei2019mean,javanmard2019analysis}
approximates the ensemble of weights by an asymptotic limit as the number of hidden units tends to infinity, so that the dynamics of SGD can be studied via certain partial different equations.

More specifically, let $\hat f(\xx; \btheta) = N^{-1} \sum_{i=1}^N \sigma(\btheta_i^\top \xx)$ be a function given by a one-hidden-layer neural net with $N$ hidden units, where $\sigma(\cdot)$ is the ReLU activation function and parameters $\btheta \triangleq [\btheta_1,\ldots,\btheta_N]^\top \in \R^{N \times d}$ are suitably randomly initialized. Consider the regression setting where we want to minimize the population risk $R_N(\btheta) =  \E[(y - \hat f(\xx; \btheta))^2]$ over parameters $\btheta$. A key observation is that this population risk depends on the parameters $\btheta$ only through its empirical distribution, i.e., $\hat \rho^{(N)} = N^{-1} \sum_{i=1}^N \delta_{\btheta_i}$ where $\delta_{\btheta_i}$ is a point mass at $\btheta_i$. This motivates us to view express $R_N(\btheta)$ equivalently as $R(\hat \rho^{(N)})$, where $R(\cdot)$ is a functional that maps distributions to real numbers. Running SGD for $R_N(\cdot)$---in a suitable scaling limit---results in a gradient flow on the space of distributions endowed with the Wasserstein metric that minimizes $R(\cdot)$. It turns out that the  empirical distribution $\hat \rho^{(N)}_k$ of the parameters after $k$ steps of SGD is well approximated by the gradient follow, as long as the the neural net is over-parametrized (i.e., $N\gg d$) and the number of steps is not too large. In particular, \cite{mei2018mean} have shown that under certain regularity conditions,
\[
\sup_{k\in[0,T/\varepsilon]\cap\mathbb{N}}\left| R(\hat \rho^{(N)})-R\left(\rho_{k\varepsilon}\right)\right|\lesssim e^{T}\sqrt{\frac{1}{N}\vee\varepsilon}\cdot\sqrt{d+\log\frac{N}{\varepsilon}},
\]
where $\varepsilon >0$ is an proxy for the step size of SGD and $\rho_{k\varepsilon}$ is the distribution of the gradient flow at time $k\varepsilon$.  In words, the out-of-sample error under $\btheta^{k}$ generated by SGD is well-approximated by that of $\rho_{k\varepsilon}$.
Viewing the optimization problem from the distributional aspect greatly simplifies the problem conceptually, as the complicated optimization problem is now passed into its limit version---for this reason, this analytical approach is called the mean field perspective. In particular, \cite{mei2018mean} further demonstrated that in some simple settings, the out-of-sample error $R(\rho_{k\varepsilon})$ of the distributional limit can be fully characterized. Nevertheless, how well does $R(\rho_{k\varepsilon})$ perform and how fast it converges remain largely open for general problems.



\subsubsection{Stability} A second way to understand the generalization ability of deep learning is through the \emph{stability} of SGD. An algorithm is considered
stable if a slight change of the input does not alter the output much. It has long been observed that a stable algorithm has a small generalization gap; examples include $k$ nearest neighbors~\citep{rogers1978finite, devroye1979distribution}, bagging~\citep{breiman1996bagging, breiman1996heuristics}, etc. The precise connection between stability and generalization gap is stated by~\cite{bousquet2002stability, shalev2010learnability}. In what follows, we formalize the idea of \emph{stability} and its connection with the generalization
gap. Let $\mathcal{A}$ denote an algorithm (possibly randomized) which takes a sample $S\triangleq\{(y_{i},\bm{x}_{i})\}_{1 \leq i \leq n}$
of size $n$ and returns an estimated parameter $\hat{\bm{\theta}}\triangleq\mathcal{A}(S)$.
Following \cite{hardt2015train}, we have the following definition
for \emph{stability}.

\begin{definition}An algorithm (possibly randomized) $\mathcal{A}$
is $\varepsilon$-uniformly stable with respect to the loss function
$\mathcal{L}(\cdot,\cdot)$ if for all datasets $S,S'$ of size $n$ which differ
in at most one example, one has 
\[
\sup_{\bm{x},y}\mathbb{E}_{\mathcal{A}}\left[\mathcal{L}\left(f(\bm{x};\mathcal{A}\left(S\right)),y\right)-\mathcal{L}\left(f(\bm{x}; \mathcal{A}\left(S'\right)),y\right)\right]\leq\varepsilon.
\]
Here the expectation is taken w.r.t.~the randomness in the algorithm
$\mathcal{A}$ and $\varepsilon$ might depend on $n$. The loss function $\mathcal{L}(\cdot,\cdot)$ takes an example (say $(\bm{x},y)$) and the estimated
parameter (say $\mathcal{A}(S)$) as inputs and outputs a real value. \end{definition}

Surprisingly, an $\varepsilon$-uniformly stable algorithm incurs
small generalization gap \emph{in expectation}, which is stated in the following
lemma.
\begin{lem}[Theorem 2.2 in \citealp{hardt2015train}]\label{lemma:stability-generalization}Let
$\mathcal{A}$ be $\varepsilon$-uniformly stable. Then the expected
generalization gap is no larger than $\varepsilon$, i.e.,
\begin{equation}
\left|\mathbb{E}_{\mathcal{A},S}\left[\frac{1}{n}\sum_{i=1}^{n}\mathcal{L}(f(\bm{x}_{i};\mathcal{A}\left(S\right)),y_{i})-\mathbb{E}_{(\bm{x},y)\sim\mathcal{D}}\left[\mathcal{L}\left(f(\bm{x};\mathcal{A}\left(S\right)),y\right)\right]\right]\right|\leq\varepsilon.\label{eq:stability-generalization-gap}
\end{equation}
\end{lem}

With Lemma~\ref{lemma:stability-generalization} in hand, it suffices to prove stability bound on specific
algorithms. It turns out that SGD introduced
in Section~\ref{sec:opt} is uniformly stable when solving smooth
nonconvex functions.

\begin{thm}[Theorem 3.12 in~\cite{hardt2015train}]\label{thm:sgd-stability}Assume
that for any fixed $(y, \bm{x})$, the loss function $\mathcal{L}(f(x;\btheta), y)$, viewed as a function of $\btheta$,
is $L$-Lipschitz and $\beta$-smooth. Consider running SGD on the
empirical loss function with decaying step size $\alpha_{t}\leq c/t$,
where $c$ is some small absolute constant. Then SGD is uniformly
stable with
\[
\varepsilon\lesssim\frac{T^{1-\frac{1}{\beta c+1}}}{n},
\]
where we have ignored the dependency on $\beta,c$ and $L$. \end{thm}Theorem~\ref{thm:sgd-stability}
reveals that SGD operating on nonconvex loss functions is indeed uniformly
stable as long as the number of steps $T$ is not large compared with
$n$. This together with Lemma~\ref{lemma:stability-generalization}
demonstrates the generalization ability of SGD in expectation. 
Nevertheless, two important limitations are worth mentioning.  First, Lemma~\ref{lemma:stability-generalization} provides an upper bound on the out-of-sample error \emph{in expectation}, but ideally, instead of an on-average guarantee under $\mathbb{E}_{\mathcal{A},S}$, we would like to have a high probability guarantee as in the convex case~\citep{feldman2019high}.
Second, controlling the generalization gap alone is not enough to achieve a small out-of-sample error, since it is unclear whether SGD can achieve a small training error within $T$ steps.

\subsubsection{Implicit regularization} 

In the presence of over-parametrization (number of parameters larger than the sample size), conventional wisdom informs us that we should apply some regularization techniques (e.g., $\ell_1\,/\, \ell_2$ regularization) so that the model will not overfit the data. However, in practice, neural networks without explicit regularization generalize well. This phenomenon motivates researchers to look at the regularization effects introduced by training algorithms (e.g., SGD) in this over-parametrized regime. While there might exits multiple, if not infinite global minima of the empirical loss~(\ref{eq:ERM_for_DL}), it is possible that practical algorithms tend to converge to solutions with better generalization powers.

Take the underdetermined linear system $\bm{X}\bm{\theta}=\bm{y}$
as a starting point. Here $\bm{X}\in\mathbb{R}^{n\times p}$ and $\bm{\theta}\in\mathbb{R}^{p}$
with $p$ much larger than $n$. Running gradient descent on the loss
$\frac{1}{2}\|\bm{X}\bm{\theta}-\bm{y}\|_{2}^{2}$ from the origin
(i.e., $\bm{\theta}^{0}=\bm{0}$) results in the solution with the minimum Euclidean
norm, that is GD converges to
\begin{align*}
\min_{\bm{\theta}\in\mathbb{R}^{p}} & \quad\|\bm{\theta}\|_{2}\qquad\text{subject to}\quad\bm{X}\bm{\theta}=\bm{y}.
\end{align*}
In words, without any $\ell_{2}$ regularization in the loss function,
gradient descent automatically finds the solution with the least $\ell_{2}$ norm.
This phenomenon, often called as \emph{implicit regularization}, not
only has been empirically observed in training neural networks, but
also has been theoretically understood in some simplified cases, e.g., logistic regression with separable data.
In logistic regression, given a training
set $\{(y_{i},\bm{x}_{i})\}_{1\leq i \leq n}$ with $\bm{x}_{i}\in\mathbb{R}^{p}$
and $y_{i}\in\{1,-1\}$, one aims to fit a logistic regression model
by solving the following program:
\begin{equation}
\min_{\bm{\theta}\in\mathbb{R}^{p}}\qquad\frac{1}{n}\sum_{i=1}^{n}\ell\big(y_{i}\bm{x}_{i}^\top \bm{\theta}^{t}\big).\label{eq:loss-logistic}
\end{equation}
Here, $\ell(u)\triangleq\log(1+e^{-u})$ denotes the logistic loss. Further
assume that the data is separable, i.e., there exists $\bm{\theta}^{*}\in\mathbb{R}^{p}$
such that $y_{i}\bm{\theta}^{*\top}\bm{x}_{i}>0$ for all $i$. Under this condition,
the loss function (\ref{eq:loss-logistic}) can be arbitrarily close to zero for certain $\btheta$ with $\| \btheta \|_2 \to \infty$.
What happens when we minimize (\ref{eq:loss-logistic}) using gradient
descent? \cite{soudry2018implicit} uncovers a striking phenomenon.

\begin{thm}[Theorem 3 in \citealp{soudry2018implicit}]Consider
the logistic regression (\ref{eq:loss-logistic}) with separable data.
If we run GD
\[
\bm{\theta}^{t+1}=\bm{\theta}^{t}-\eta\frac{1}{n}\sum_{i=1}^{n}y_{i}\bm{x}_{i}\ell'\big(y_{i}\bm{x}_{i}^\top \bm{\theta}^{t}\big)
\]
from any initialization $\bm{\theta}^{0}$ with appropriate step size
$\eta>0$, 
then normalized $\bm{\theta}^{t}$ converges to a solution with the maximum $\ell_2$ margin. That is,
\begin{equation}
\lim_{t\to\infty}\frac{\bm{\theta}^{t}}{\|\bm{\theta}^{t}\|_{2}}=\hat{\bm{\theta}},\label{eq:converge-in-direction}
\end{equation}
where $\hat{\bm{\theta}}$ is the solution to the hard margin support
vector machine: 
\begin{equation}
\hat{\bm{\theta}}\triangleq\arg\min_{\bm{\theta}\in\mathbb{R}^{p}}\|\bm{\theta}\|_{2},\qquad\text{subject to}\quad y_{i}\bm{x}_{i}^\top \bm{\theta} \geq1\quad\text{for all }1\leq i\leq n.\label{eq:SVM}
\end{equation}
\end{thm}

The above theorem reveals that gradient descent, when solving logistic
regression with separable data, implicitly regularizes the iterates towards
the $\ell_{2}$ max margin vector (cf.~(\ref{eq:converge-in-direction})),
without any explicit regularization as in (\ref{eq:SVM}). Similar
results have been obtained by \cite{ji2018risk}. In addition, \cite{gunasekar2018characterizing}
studied algorithms other than gradient descent and showed that coordinate descent produces a solution with the maximum $\ell_1$ margin.

Moving beyond logistic regression, which can be viewed as a one-layer neural net, the theoretical understanding of implicit regularization in deeper neural networks is still limited; see~\cite{gunasekar2018implicit} for an illustration in deep linear convolutional neural networks.



%

\section{Discussion}\label{sec:discuss}

Due to space limitations, we have omitted several important deep learning models; notable examples include deep reinforcement learning~\citep{mnih2015human}, deep probabilistic graphical models~\citep{salakhutdinov2009deep}, variational autoencoders~\citep{kingma2013auto}, transfer learning~\citep{yosinski2014transferable}, etc. Apart from the modeling aspect, interesting theories on generative adversarial networks~\citep{arora2017generalization, bai2018approximability}, recurrent neural networks~\citep{AL2018-RNNgen}, connections with kernel methods~\citep{jacot2018neural,arora2019fine} are also emerging. We have also omitted the inverse-problem view of deep learning where the data are assumed to be generated from a certain neural net and the goal is to recover the weights in the NN with as few examples as possible. Various algorithms (e.g., GD with spectral initialization) have been shown to recover the weights successfully in some simplified settings~\citep{zhong2017recovery, soltanolkotabi2017learning, goel2018learning, mondelli2018connection, chen2018gradient,fu2018local}.

In the end, we identify a few important directions for future research.
\begin{itemize}
\item{\emph{New characterization of data distributions.} The success of deep learning relies on its power of efficiently representing complex functions relevant to real data. Comparatively, classical methods often have optimal guarantee if a problem has a certain known structure, such as smoothness, sparsity, and low-rankness~\citep{stone1982optimal, donoho1994ideal, candes2009power,chen2019noisy}, but they are insufficient for complex data such as images. How to characterize the high-dimensional real data that can free us from known barriers, such as the curse of dimensionality is an interesting open question? 
}


\item \emph{Understanding various computational algorithms for deep learning.} As we have emphasized throughout this survey, computational algorithms (e.g., variants of SGD) play a vital role in the success of deep learning. They allow fast training of deep neural nets and probably contribute towards the good generalization behavior of deep learning in practice. Understanding these computational algorithms and devising better ones are crucial components in understanding deep learning.

\item{\emph{Robustness.} It has been well documented that DNNs are sensitive to small adversarial perturbations that are indistinguishable to humans~\citep{szegedy2013intriguing}. This raises serious safety issues once if deploy deep learning models in applications such as self-driving cars, healthcare, etc. It is therefore crucial  to refine current training practice to enhance robustness in a principled way~\citep{singh2018hierarchical}.  
}

\item{\emph{Low SNRs.} Arguably, for image data and audio data where the signal-to-noise ratio (SNR) is high, deep learning has achieved great success. In many other statistical problems, the SNR may be very low. For example, in financial applications, the firm characteristic and covariates may only explain a small part of the financial returns; in healthcare systems, the uncertainty of an illness may not be predicted well from a patient's medical history. How to adapt deep learning models to excel at such tasks is an interesting direction to pursue?

}

\end{itemize}


\section*{Acknowledgements}
J.~Fan is supported in part by the NSF grants DMS-1712591 and DMS-1662139, the NIH grant R01-GM072611 and the ONR grant N00014-19-1-2120. We thank Ruying Bao, Yuxin Chen, Chenxi Liu, Weijie Su, Qingcan Wang and Pengkun Yang for helpful comments and discussions.

\bibliographystyle{plain}
\bibliography{bibDeepLearning}

\end{document}